%% file: main.tex
\documentclass{article}

\usepackage[accepted]{icml2020}

\usepackage[utf8]{inputenc} 
\usepackage[T1]{fontenc}    
\usepackage{hyperref}       
\usepackage{url}            
\usepackage{booktabs}       
\usepackage{amsfonts}       
\usepackage{nicefrac}       
\usepackage{microtype}      
\usepackage{xcolor}
\usepackage{amsmath}
\usepackage{amssymb}
\usepackage{mathabx}
\usepackage{graphicx}
\usepackage{subfig}
\usepackage{float}

\DeclareMathOperator*{\argmin}{arg\,min}
\newcommand\numberthis{\addtocounter{equation}{1}\tag{\theequation}}
\usepackage{bbm}

\icmltitlerunning{Self-supervised Learning of Distance Functions for Goal-Conditioned Reinforcement Learning}

\begin{document}

\twocolumn[
\icmltitle{Self-supervised Learning of Distance Functions for Goal-Conditioned Reinforcement Learning}

\icmlsetsymbol{equal}{*}

\begin{icmlauthorlist}
\icmlauthor{Srinivas Venkattaramanujam}{mila,mcgill}
\icmlauthor{Eric Crawford}{mila,mcgill}
\icmlauthor{Thang Doan}{mila,mcgill}
\icmlauthor{Doina Precup}{mila,mcgill}
\end{icmlauthorlist}

\icmlaffiliation{mila}{Mila, Montreal, Canada}
\icmlaffiliation{mcgill}{McGill University, Montreal, Canada}

\icmlcorrespondingauthor{Srinivas Venkattaramanujam}{sri.venkattaramanujam@mail.mcgill.ca}

\icmlkeywords{Machine Learning, ICML}

\vskip 0.3in
]

\printAffiliationsAndNotice{}

\begin{abstract}
 A crucial requirement of goal-conditioned policies is to be able to determine whether the goal has been achieved. Having a notion of distance to a goal is thus a crucial component of this approach. However, it is not straightforward to come up with an appropriate distance, and in some tasks, the goal space may not even be known a priori. In this work we learn, in a self-supervised manner, a distance-to-goal estimate which is computed in terms of the average number of actions that would need to be carried out to reach the goal. In order to learn the distance estimate, we propose to learn an embedding space such that the distance between points in this space corresponds to the square-root of the average number of timesteps required to go from the first state to the second and back, i.e. the commute time between the states. We discuss why such an embedding space is guaranteed to exist and provide a practical method to approximate it in the online reinforcement learning setting. Experimental results in a number of challenging domains demonstrate that our approach can greatly reduce the amount of domain knowledge required by existing algorithms for goal-conditioned reinforcement learning.

\end{abstract}

\section{Introduction}
\input{icml2020/sections/introduction.tex}

\section{Related Work}
\input{icml2020/sections/related.tex}

\section{Background}
\input{icml2020/sections/background.tex}

\section{Method}
\input{icml2020/sections/algo.tex}

\section{Experimental Results}
\input{icml2020/sections/experiments.tex}

\section{Conclusion}
\input{icml2020/sections/conclusion.tex}



\bibliography{bibliography}
\bibliographystyle{icml2020}
\clearpage
\input{icml2020/sections/appendix.tex}
\end{document}

%% file: icml2020/sections/introduction.tex

Reinforcement Learning (RL) is a framework for training agents to interact optimally with an environment.
Recent advances in RL have led to algorithms that are capable of succeeding in a variety of environments, ranging from video games with high-dimensional image observations \cite{DBLP:journals/corr/MnihKSGAWR13, mnih2015humanlevel} to continuous control in complex robotic tasks \cite{DDPG,TRPO}. Meanwhile, innovations in training powerful function approximators have all but removed the need for hand-crafted state representation, thus enabling RL methods to work with minimal human oversight or domain knowledge. However, one component of the RL workflow that still requires significant human input is the design of the reward function that the agent optimizes.

One way of alleviating this reliance on human input is by allowing the agent to condition its behavior on a provided goal \cite{Kaelbling93b, schaul2015universal}, and training the agent to achieve (some approximation of) all possible goals afforded by the environment. A number of algorithms have recently been proposed along these lines, often making use of curriculum learning techniques to discover goals and train agents to achieve them in a structured way \cite{IJCAI17-Narvekar, Florensa2018AutomaticGG}. At the end of this process, the agent is expected to be able to achieve any desired goal.

An important component of this class of algorithm is a distance function, used to determine whether the agent has reached its goal;  this can also require  human input and domain knowledge.
In past work, it has been common to assume that the goal space is known and use the $L2$ distance between the current state and the goal. However, this straightforward choice is not satisfactory for general environments, as it does not take environment dynamics into account. For example, it is possible for a state to be close to a goal in terms of $L_2$ distance, and yet be far from satisfying it in terms of environment dynamics.


We propose a self-supervised method for learning a distance between a state and a goal which accurately reflects the dynamics of the environment. We begin by defining the distance between two states as the square root of the average number of time steps required to move from the first state to the second and back for the first time under some policy $\pi$. To make this distance usable as part of a goal-conditioned reward function, we train a neural network to approximate this quantity from data. The distance network is trained online, in conjunction with the training of the goal-conditioned policy.

The contributions of this work are as follows.
i) We propose a self-supervised approach to learn a distance function, by learning an embedding with the property that the $p$-norm between the embeddings of two states approximates the average temporal distance between the states according to a policy $\pi$,
ii) we show that our method is approximating a theoretically motivated quantity and discuss the connection between our approach and the graph Laplacian,
iii) we demonstrate that the learned distance estimate can be used in the online setting in goal-conditioned policies,
iv) we develop an automatic curriculum generation mechanism that takes advantage of our distance learning algorithm, and v) we explain a phenomenon that arises due to learning the distance function using samples from the behavior policy. Our method solves complex tasks without prior \textit{domain knowledge} in the online setting in three different scenarios in the context of goal-conditioned policies - a) the goal space is the same as the state space, b) the goal space is given but an appropriate distance is unknown and c) the state space is accessible, but only a subset of the state space represents desired goals, and this subset is known a priori.

%% file: icml2020/sections/related.tex
Goal-conditioned RL aims to train agents that can reach any goal provided to them. Automatic goal generation approaches such as \cite{Florensa2018AutomaticGG, florensa2017reverse} focus on automatically generating goals of appropriate difficulty for the agent, thereby facilitating efficient learning. These methods utilize domain knowledge to define a goal space and use $L2$ distance as the distance function in the goal space. However, in most tasks the goal space is inaccessible or an appropriate distance function in the goal space is unknown. There have been recent efforts on learning an embedding space for goals in an unsupervised fashion using the reconstruction error and the $L2$ distance is then computed in the learned embedding space
\cite{pere2018unsupervised, Imagined_goals,goal_embedding}. The main drawback of these approaches is that they do not capture the environment dynamics.

\cite{andrychowicz2017hindsight} and \cite{rauber2018hindsight} focus on improving the sample efficiency of the goal-conditioned policies by relabeling or reweighting the reward from a goal on which the trajectory was conditioned to a different goal that was a part of the trajectory. Our method is complementary to these approaches since they rely on prior knowledge of the goal space and use the $L1$ or $L2$ distance in the goal space to determine whether the goal has been reached.

Similar to our work, \cite{savinov2018semiparametric} and \cite{savinov2018episodic} trained a network $R$ to predict whether the distance in actions between two states is smaller than some fixed hyperparameter $k$. However, \cite{savinov2018semiparametric} and \cite{savinov2018episodic} were done in the context of supervised learning-based navigation and intrinsic motivation, respectively, in contrast to our work. \cite{savinov2018semiparametric} proposed a non-parametric graph based memory module for navigation where the nodes correspond to landmarks in the environment and the nodes judged similar by the network $R$ are connected by an edge; goal-oriented navigation is performed by using a locomotion network $L$ trained using supervised learning to reach intermediate way-points selected as a result of localization and planning on the learned graph. \cite{savinov2018episodic} used the network to to provide agents with an exploration bonus for visiting novel states; given a state $s$ visited by the agent, an exploration bonus was provided if the network judged $s$ to be far from the states in a buffer storing a representative sample of states previously visited by the agent. 

\cite{ghosh2018learning} defines the actionable distance between states $s_1$ and $s_2$ in terms of expected Jensen-Shannon Divergence between $\pi(a|s, s_1)$ and $\pi(a|s, s_2)$, where $\pi(a|s, g)$ is a fully trained goal-conditioned policy. They then train an embedding such that the $L2$ distance between the embeddings of $s_1$ and $s_2$ is equal to the actionable distance between $s_1$ and $s_2$.
This differs from our approach in that we use a different objective for training the distance function, and, more importantly, we do not assume availability of a pre-trained goal-conditioned policy; rather, in our work the distance function is trained online, in conjunction with the policy.

\cite{wu2018the} learns a state representation using the eigenvectors of the Laplacian of the graph induced by a fixed policy and demonstrates its suitability to reward shaping in sparse reward problems. Our method aims to learn an embedding of the states without computing the eigenvectors of the graph Laplacian. However, the justification of our approach relies on why eigenvectors of the Laplacian are insufficient when the distance between the states in the embedding space is crucial. We discuss the details of this and the connection to the commute time in sections \ref{sec:existence_of_emb_space} and \ref{sec:ectd}. Furthermore, our approach differs by not using negative sampling; only the information present within trajectories are used to obtain the embeddings.

%% file: icml2020/sections/background.tex

\subsection{Goal-Conditioned Reinforcement Learning}
In the standard RL framework, the agent is trained to solve a single task, specified by the reward function. Goal-conditioned reinforcement learning generalizes this to allow agents capable of solving multiple tasks \cite{schaul2015universal}. We assume a goal space $G$, which may be identical to the state space or related to it in some other way, and introduce the goal-augmented state space $S_G = S \bigtimes G$.
Given some goal $g \in G$, the policy $\pi(a_t | s_t, g)$, reward function $r(s_t, g, a_t)$ and value function $V_{\pi}(s_t, g)$ are conditioned on the goal $g$ in addition to the current state. The objective is to train the agent to achieve all goals afforded by the environment.

We assume the goal space is either identical to or a subspace of the state space, that all trajectories begin from a single start state $s_0$, and that the environment does not provide a means of sampling over all possible goals (instead, goals must be discovered through experience). Moreover, we require a distance function $d(s, g)$; agents are given a reward of 0 at all timesteps until $d(s, g) < \epsilon$, for hyperparameter $\epsilon \in \mathbb{R}^+$, at which point a reward of 1 is provided and the episode terminates.



\subsection{Goal Generation and Curriculum Learning}
In order to train agents to achieve all goals in this setting, it is desirable to have a way of systematically exploring the state space in order to discover as many goals as possible, as well as a means of tailoring the difficulty of goals to the current abilities of the agent (a form of goal-based curriculum learning). An algorithmic framework satisfying both of these requirements was proposed in \cite{Florensa2018AutomaticGG}. Under this framework, one maintains a working set of goals, and alternates between two phases. In the \textit{policy-learning} phase, the agent is trained (using an off-the-shelf RL algorithm) to achieve goals sampled uniformly from the working set. In the \textit{goal-selection} phase, the working set of goals is adapted to the current abilities of the agent in order to enable efficient learning in the next policy-learning stage.
In particular, the aim is to have the working set consist of goals that are of intermediate difficulty for the agent; goals that are too hard yield little reward to learn from, while goals that are too easy leave little room for improvement. Formally, given hyperparameters $R_{min}, R_{max} \in (0, 1)$, a goal $g$ is considered to be a \textit{Goal of Intermediate Difficulty} (GOID) if $R_{min} < V_{\pi_\theta}(s_0, g) < R_{max}$, where $V_{\pi_\theta}(s_0, g)$ is the undiscounted return.

\subsection{Multidimensional scaling}
The study of multidimensional scaling (MDS) is concerned with finding a low-dimensional configuration of objects by taking as input a set of pairwise dissimilarities between the objects \cite{modMDS}. The resulting low-dimensional configuration has to be such that the distance in this the low-dimensional configuration between any pair of objects best preserves the corresponding pairwise dissimilarities provided as input. As the dissimilarities are mapped to distances in the low-dimensional space, the input dissimilarities cannot be arbitrary. They must be symmetric, non-negative and obey the triangle inequality.  The discussion and notation used in this section follows \cite{modMDS}.

\subsubsection{Classical MDS}\label{sec:cMDS}
MDS in its original form is now called Classical MDS (cMDS) or Toegerson scaling. Classical MDS assumes that the dissimilarities are distances in some Euclidean space. The embeddings produced by cMDS preserve the input distances exactly whenever the inputs are Euclidean distances. 

Provided with a matrix of pairwise distances $D$, cMDS proceeds as follows. First a matrix $D^{(2)}$ of squared pairwise distances is formed. Then a matrix $B$ is obtained by double centering $D^{(2)}$, i.e $B=-\frac{1}{2}JD^{(2)}J$ where $J=I - \frac{1}{n}\mathbf{1}\mathbf{1}^{T}$ and $\mathbf{1}$ is a vector of all ones. $B$ is symmetric and positive-semidefinite (details in Appendix \ref{sec:cmds_appendix}). Finally, an eigen-decomposition on $B$ produces $B=Q \Lambda Q^{T}$, where $\Lambda$ is a diagonal matrix whose elements are the eigenvalues of $B$ arranged in descending order and the columns of $Q$ are the corresponding eigenvectors. An embedding $X^{'}$ that preserves the Euclidean distance is then obtained using $X^{'}=Q\Lambda ^{\frac{1}{2}}$.
\subsubsection{Metric MDS}
\label{sec:mMDS}
Let $\delta_{ij}$ denote the dissimilarity between objects $i$ and $j$, and let $d_{ij}(X)$ denote a distance metric between the $i^{th}$ and $j^{th}$ rows of $X$ denoted by $x_i$ and $x_j$ respectively. Typically, the distance metric $d$ is the Euclidean distance. $x_k$ is the representation of the object $k$ provided by $X$.
MDS minimizes a quantity called \textit{stress}, denoted by $\sigma_r(X)$, defined as 
\begin{equation}
    \label{eq:stress}
    \sigma_r(X)=\sum_{i<j} w_{ij} (d_{ij}(X) - \delta_{ij})^{2}
\end{equation}
where $w_{ij} \geq 0$, and $w_{ij} = w_{ji}$. Any meaningful choice of weights that satisfies these constraints can be used.

In equation (\ref{eq:stress}), $\delta_{ij}$ can be replaced by $f(\delta_{ij})$. If $f$ is continuous, the approach is then called metric MDS. A generalization of the stress is defined as 
\begin{equation}
    \label{eq:stress_gen}
    \sigma_G(X)=\sum_{i<j} w_{ij} (f(d_{ij}(X)) - f(\delta_{ij}))^{2}
\end{equation}
where $f(x)=x$ corresponds to the raw stress $\sigma_r$. 
In general, metric MDS does not admit an analytical solution.
Instead, it is solved iteratively, and convergence to a global minimum is not guaranteed.

\subsection{Spectral Embeddings of Graphs}
\label{sec:spectral_graph_drawing}
Given a simple, weighted, undirected and connected graph $G$, the Laplacian of the graph is defined as $L=D-W$ where $W$ is the weight matrix and $D$ is the degree matrix. The eigenvectors corresponding to the smallest eigenvalues of the graph Laplacian are used to obtain an embedding for the nodes and have been shown useful in several applications such as spectral clustering \cite{tutospecclustr} and spectral graph drawing \cite{Koren:2003:SGD:1756869.1756936}. In spectral embedding methods, a $k$-dimensional embedding of the node $i$ is obtained by taking the $i^{th}$ components of the $k$ eigenvectors corresponding to the $k$ smallest non-zero eigenvalues. 

\subsection{Markov Chain based distances}
The average first passage time from state $i$ to $j$ is defined as the expected number of steps to reach $j$ for the first time after starting from $i$. We write the average first passage time $m(j|i)$ recursively as 
\begin{equation*}
    m(j|i) = \begin{cases}
                0 & \text{if } i = j\\
                1 + \sum_{k \in S} P(k|i) m(k|i) & \text{if } i \neq j \\   
             \end{cases}
\end{equation*}
A related quantity, the average commute time $n(i, j)$ is defined as $n(i,j)=m(i|j) + m(j|i)$. Average commute time is a distance metric as noted in \cite{Fouss:2005:NWC:1092358.1092536}.

%% file: icml2020/sections/algo.tex
In this section we introduce
an action-based distance measure for use in trajectory-based reinforcement learning which captures the dynamics of the environment. We then present a method for automatically learning an estimator of that distance using samples generated by a policy $\pi$.

\subsection{Learned Action Distance}
\label{sec:distance}
We propose to learn a task-specific distance function where the distance between states $s_1$ and $s_2$ is defined as half of the commute-time, which we call the \textit{action distance}.  Defining the distance in terms of reachability of the states captures the environment dynamics as experienced by the agent under the policy $\pi$. In order to learn a distance estimator we propose to learn an embedding such that the distance between the embeddings of a pair of states is equal to the action distance between the states.
Formally, let $s_i, s_j \in \mathcal{S}$, and define the action distance $d^\pi(s_i, s_j)$ as:
\begin{align}
    d^{\pi}(s_i, s_j) = \frac{1}{2} m(s_j|s_i) + \frac{1}{2} m(s_i|s_j)
    \label{eq:ad_commute_time}
\end{align}

In general, $d^\pi(s_i, s_j)$ is difficult to compute; this is problematic, since this distance is intended to be used for detecting when goals have been achieved and will be called frequently. Therefore, we propose to train a neural network to estimate it. Specifically, we learn an embedding function $e_\theta$ of the state space, parameterized by vector $\theta$, such that the $p$-norm between a pair of state embeddings is close to the action distance between the corresponding states. The objective function used to train $e_\theta$ is:
\begin{equation}
\label{eq:1}
    \theta^* = \argmin_\theta (||e_\theta(s_i) - e_\theta(s_j)||_p^q - d^{\pi}(s_i, s_j))^2
\end{equation}

In general, ensuring $d^{\pi}(s_i, s_j)$ is computed using equal proportion of $m(s_j|s_i)$ and $m(s_i|s_j)$ leads to practical difficulties. Hence, due to practical considerations, we redefine action distance to 
\begin{align*}
    d^{\pi}(s_i, s_j) = \frac{\rho(s_i)m(s_j|s_i)}{\rho(s_i) + \rho(s_j)} + \frac{\rho(s_j) m(s_i|s_j)}{\rho(s_i) + \rho(s_j)}
    \label{eq:action_distance}
\end{align*}

where $\rho$ is the stationary distribution of the Markov chain (we assume the stationary distribution exists) and $m(\cdot | \cdot)$ is estimated from the trajectories as follows

\begin{align}
    m(s_j | s_i) = E_{\tau \sim \pi , t \sim \overline{t}(s_i, \tau), t^{'} = \min\{ \overline{t}(s_j, \tau )\geq m\}} \left [|t - t' |\right ]
\end{align}

where $\overline {t}(s, \tau)$ is a uniform distribution over all temporal indices of $s$ in $\tau$ and the expectation is taken over trajectories $\tau$ sampled from $\pi$ such that $s_i$ and $s_j$ both occur in $\tau$. If $\pi$ is a goal-conditioned policy we also average over goals $g$ provided to $\pi$. In the next section we discuss the existence of the embedding space that preserves action distance (equation \ref{eq:ad_commute_time}), its connection to the graph Laplacian, and a practical approach to approximate it.



When used as part of an algorithm to learn a goal-conditioned policy, the distance predictor can be trained using the trajectories collected by the behavior policy during the policy-learning phase. We call this case the \textit{on-policy} distance predictor. We emphasize that the on-policy nature of this distance predictor is independent of whether the policies or value functions are learned on-policy.
While such an on-policy distance possesses desirable properties, such as a simple training scheme, it also has drawbacks. Both the behavior policy and goal distribution will change over time and thus the distance function will be non-stationary.
This can create a number of issues, the most important of which is difficulty in setting the threshold $\epsilon$. Recall that in our setting, the goal is considered to be achieved and the episode terminated once $d^\pi(s, g) < \epsilon$, where $\epsilon$ is a threshold hyperparameter.
This thresholding creates an $\epsilon$-sphere around the goal, with the episode terminating whenever the agent enters this sphere. The interaction between the $\epsilon$-sphere and the non-stationarity of the on-policy distance function causes a subtle issue that we dub the \textit{expanding $\epsilon$-sphere} phenomenon, discussed in detail in Section \ref{sec:expanding_sphere}.

An alternative approach to learning the distance function from the trajectories generated by the behavior policy is to apply a random policy for a fixed number of timesteps at the end of each episode. The states visited under the random policy are then used to train the distance function.
Since the random policy is independent of the behavior policy, we describe the learned distance function as \textit{off-policy} in this case. The stationarity of the random policy helps in overcoming the expanding $\epsilon$-sphere phenomenon of the on-policy distance predictor.

\subsection{Existence and Approximation of the Embedding Space \label{sec:existence_of_emb_space}}
Our approach relies on spectral theory of graphs to obtain a representation of states, similar to \cite{wu2018the}. We note that the Laplacian $L=D-W$ of \cite{wu2018the} in the finite state setting is given by $W_{uv}=\frac{1}{2}\rho(u) P^\pi(v|u) + \frac{1}{2} \rho(v) P^\pi(u|v)$ and the transition probabilities $P_{uv}=\frac{1}{2} P^\pi(v|u) + \frac{1}{2} \frac{\rho (v)}{\rho (u)}P^\pi(u|v)$ (details in \ref{sec:spectral_graph_drawing}). The random walk on undirected weighted graphs defines a Markov chain and hence, Markov chain based distances are useful to define dissimilarities between nodes. Specifically, \cite{Fouss:2005:NWC:1092358.1092536} define the similarity of nodes in weighted undirected graphs using commute times and show the  existence of a Euclidean embedding space where the distance between embeddings of nodes $i$ and $j$ correspond to the square root of the average commute time between them, $\sqrt{n(i,j)}$. Such a Euclidean space can be computed using the pseudo-inverse of the graph Laplacian $L^{\dagger}$, and the representation of a node $i$ is the $i^{th}$ row of $Q\Lambda^{\frac{1}{2}}$ where the columns of $Q$ are the eigenvectors of $L^{\dagger}$ and $\Lambda$ is a diagonal matrix of the corresponding eigenvalues. This is also the solution given by cMDS when $D^{(2)}(X)_{ij}=n(i,j)$ or when $B_{ij}=L^\dagger_{ij}$. Further discussion is provided in \ref{sec:ectd}.

Even though the embedding space with the desired properties exists, neither the matrix of pairwise commute times nor $L^{\dagger}$ are available in the RL setting and hence cMDS cannot be applied. Hence, we approximate the embedding space using metric MDS (equation \ref{eq:stress_gen}) where $\delta_{ij}=\sqrt{n(i,j)}$. Metric MDS provides flexibility in the choices of $f$ and weights $w_{ij}$. We choose $f$ to be the square function i.e $f(x)=x^2$. This choice of $f$ stems from practical considerations - it is easier to estimate the average first passage times between $m(j|i)$ and $m(i|j)$ independently using the trajectories in RL compared to estimating the commute time $n(i, j)=\sqrt{m(j|i) + m(i|j)}$. Hence, our objective function is of the form 
\begin{equation}
\label{eq:s_stress}
    \sigma_S(X)=\sum_{i<j} w_{ij} (||x_i - x_j||_2^2 - \frac{1}{2}(m(j|i) + m(i|j)))^{2}
\end{equation}
where $x_i$ is the embedding of node $i$. By noting that the quantity in equation (\ref{eq:s_stress}) is the mean squared error (MSE) and MSE computes the mean of the labels drawn from a distribution for each input, we can write equation (\ref{eq:s_stress}) as
\begin{align*}
    \sigma_S(X) &=\frac{1}{2}\sum_{i=1}^n\sum_{j=1}^n w_{ij} (||x_i - x_j||_2^2 - k(i,j))^{2}
    \numberthis
    \label{eq:stress_justification}
\end{align*}
where $k(i, j)=\mathbbm{1}m(j|i) + (1-\mathbbm{1}) m(i|j)$ and $\mathbbm{1}\sim Bern(0.5)$. Thus, sampling $m(j|i)$ and $m(i|j)$ in equal proportion provides the required averaging. In practice, however, it is simpler to sample $\mathbbm{1} \sim Bern\left(\frac{\rho(u)}{\rho(u)+\rho(v)}\right)$ than $Bern(0.5)$, motivating our definition in equation \eqref{eq:action_distance}. 

For the choice of $w_{ij}$, the discussion of \cite{wu2018the} in Appendix \ref{sec:spectral_graph_drawing} suggests $w_{ij}=\rho(i)P(j|i)$. A better choice would be consider the probabilities over all the time steps instead of the single step transition probability. Hence, we define the weight for the pair $(x_i,x_j)$ by $w_{ij}=\rho(i) + \rho(j)$, the frequency of visiting $i$ and $j$ together. In \ref{sec:approx_with_mmds}, we discuss the weights for the case when $m(.|.)$ is not available and has to be estimated using the trajectories.



To summarize, \cite{Fouss:2005:NWC:1092358.1092536} shows the existence of the embedding space where the distance between points in the embedding space corresponds to the square root of the expected commute times between nodes. Therefore, the existence of the embedding space which preserves the action distance (equation \ref{eq:ad_commute_time}) is also guaranteed. The computation of this embedding space requires quantities that are unavailable in the reinforcement learning setting. Hence, we propose to approximate the embedding space using metric MDS, and provide a set of values for the weights $w_{ij}$ that are meaningful and practically feasible to compute. In \ref{sec:q_effect} we discuss the effect of few choices of $f$ and how it can be used to control the trade-off between faithfully representing smaller and larger distances.

\subsection{Action Noise Goal Generation}\label{sec:goal_gen}
Our algorithm maintains a working set of goals for the policy to train against. The central challenge in designing a curriculum is coming up with a way to ensure that the working set contains as many goals of intermediate difficulty (GOID) as possible. The most straightforward way of generating new GOID goals from old ones is by applying perturbations to the old goals. The downside of this simple approach is that the noise must be carefully tailored to the environment of interest, which places a significant demand for domain knowledge about the nature of the environment's state space. Consider, for instance, that in an environment where the states are represented by images it would be difficult to come up with any kind of noise such that the newly generated states are feasible (i.e. in $S$). Another option is to train a generative neural network to generate new GOID goals, as proposed by GoalGAN \cite{Florensa2018AutomaticGG}; however, this introduces significant additional complexity.

A simple alternative is to employ action space noise. That is, to generate new goals from an old goal, reset the environment to the old goal and take a series of actions using a random policy; take a random subset of the encountered states as the new goals. The states generated in this way are guaranteed to be both feasible and near the agent's current ability. Moreover, applying this approach requires only knowledge of the environment's action space, which is typically required anyway in order to interact with the environment. A similar approach was used in \cite{florensa2017reverse}, but in the context of generating a curriculum of start states growing outward from a fixed goal state.

If implemented without care, action space noise has its own significant drawback: it requires the ability to arbitrarily reset the environment to a state of interest in order to start taking random actions, a strong assumption which is not satisfied for many real-world tasks. Fortunately, we can avoid this requirement as follows. Whenever a goal is successfully achieved during the policy optimization phase, rather than terminating the trajectory immediately, we instead continue for a fixed number of timesteps using the random policy. During the goal selection phase, we can take states generated in this way for GOID goals as new candidate goals. The part of the trajectory generated under the random policy is not used for policy optimization. This combines nicely with the off-policy method for training the distance predictor, as the distance predictor can be trained on these trajectories; this results in a curriculum learning procedure for goal-conditioned policies that requires minimal domain knowledge.

%% file: icml2020/sections/experiments.tex




As a test bed we use a set of 3 Mujoco environments in which agents control simulated robots with continuous state/action spaces and complex dynamics. The first environment is called Point Mass, wherein an agent controls a sphere constrained to a 2-dimensional horizontal plane which moves through a figure eight shaped room. The state space is 4-dimensional, specifying position and velocity in each direction, while the 2-dimensional action space governs the sphere's acceleration. In the other two environments, the agent controls a quadripedal robot with two joints per leg, vaguely resembling an ant. The 41-dimensional state space includes the center-of-mass of the Ant's torso as well as the angle and angular velocity of the joints, while the action space controls torques for the joints. The Ant is significantly more difficult than point mass from a control perspective, since a complex gait must be learned in order to navigate through space. We experiment with this robot in two different room layouts: a simple rectangular room (Free Ant) and a U-shaped maze (Maze Ant). Further discussion of our environments can be found in \cite{Florensa2018AutomaticGG} and \cite{pmlr-v48-duan16}. For certain experiments in the Maze Ant environment, the canonical state space is replaced by a pixel-based state space giving the output of a camera looking down on the maze from above.

In the first set of experiments we seek to determine whether our online distance learning approach can replace the hand-coded distance function in the GoalGAN algorithm, thereby eliminating the need for a human to choose and/or design the distance function for each new environment. We experiment with different choices for the goal space, beginning with the simplest case in which the goal space is the $(x, y)$ coordinates of the robot's center-of-mass (i.e. a subspace of the state space) before proceeding to the more difficult case in which the goal and state spaces are identical. Next, we present the results of training the distance predictor using pixel inputs in the batch setting in the Ant Maze environment, showing that our method can learn meaningful distance functions from complex observations. Next, we empirically demonstrate the expanding $\epsilon$-sphere phenomenon mentioned in Section \ref{sec:distance}, which results from training the distance predictor with on-policy samples. Finally, we show that the goal generation approach proposed in Appendix \ref{sec:goal_gen} yields performance that is on par with GoalGAN while requiring significantly less domain knowledge. In Appendix \ref{sec:additional_experiments}, we present additional experiments on comparing spectral graph drawing with commute-time preserving embeddings in a tabular setting and discuss the effects of some hyperparameters.

\subsection{GoalGAN with Learned Action Distance}
Here we test whether our proposed method can be used to learn a distance function for use in GoalGAN, in place of the hard-coded L2 distance. We explore two methods for learning the distance function: 1) the on-policy approach, in which the distance function is trained using states from the trajectories sampled during GoalGAN's policy-learning phase, and 2) the off-policy approach, in which the distance function is trained on states from random trajectories sampled at the end of controlled trajectories during the policy-learning phase. For the embedding network $e$ we use a multi-layer perceptron with one hidden layer with $64$ hidden units and an embedding size of $20$. As we are interested in an algorithm's ability to learn to accomplish all goals in an environment, our evaluation measure is a quantity called \textit{coverage}: the probability of goal completion, averaged over all goals in the environment. For evaluation purposes, goal-completion is determined using Euclidean distance and a threshold of $\epsilon = 0.3$ for Point Mass and $\epsilon = 1.0$ for Ant. Since the goal spaces are large and real-valued, in practice we approximate coverage by partitioning the maze into a fine grid and average over goals placed at grid cell centers. Completion probability for an individual goal is taken as an empirical average over a small number of rollouts.

\subsubsection{XY Goal Space}
\begin{figure}[h!]
\centering
\includegraphics[width=0.49\textwidth]{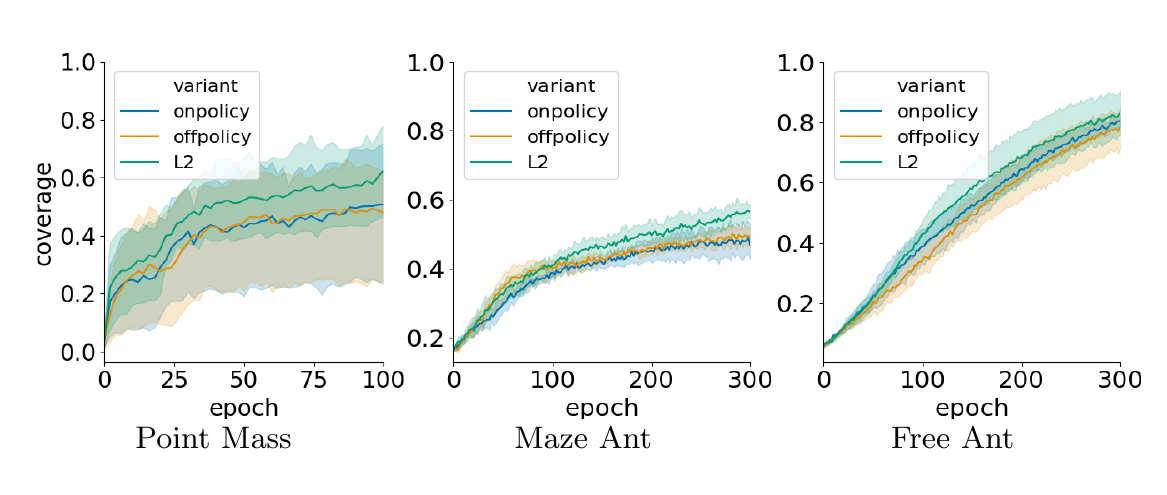}
\caption{Coverage plots for the $(x, y)$ goal space. Our method does not require domain knowledge unlike the $L2$ distance.}
\label{fig:visualization_xy_goal_space}
\end{figure}

In this experiment, the goal space is the $(x, y)$ coordinates of the robot's center-of-mass (i.e. a subspace of the state space). We compare our approach with the baseline where the $L2$ distance is used as the distance metric in the goal space. In this setting we are able to achieve performance comparable to the baseline without using any domain knowledge, as shown in Figure \ref{fig:visualization_xy_goal_space}.

\subsubsection{Full State Space as Goal Space}
In this setting the goal space is the entire state space, and the objective of the agent is to learn to reach all feasible configurations of the state space. This is straightforward for the Point Mass environment, as its 4-dimensional state space is a reasonable size for a goal space while still being more difficult than the $(x, y)$ case explored in the previous section. In contrast, the Ant environment's 41-dimensional state space is quite large, making it difficult for any policy to learn to reach every state even with a perfect distance function. Consequently, we employ a \textit{stabilization} step for generating goals from states, which makes goal-conditioned policy learning tractable while preserving the difficulty for learning the distance predictor. This step is described in detail in Appendix \ref{sec:stable}.

\begin{figure}[h!]
\centering
\includegraphics[width=0.49\textwidth]{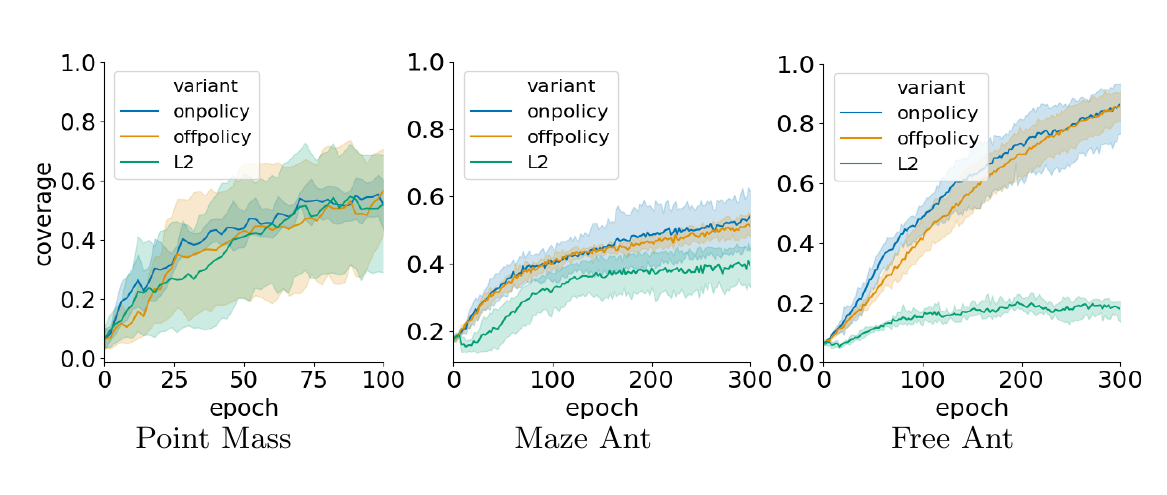}
\caption{Coverage plots for the full goal space.}
\label{fig:visualization_full_state}
\end{figure}

\begin{figure*}[!htb]
    \centering
   \begin{minipage}{\textwidth}
    \centering
    \subfloat[]{\includegraphics[trim={0.5cm 12cm 2.5cm 1cm}, clip, width=0.13\textwidth, height=2cm]{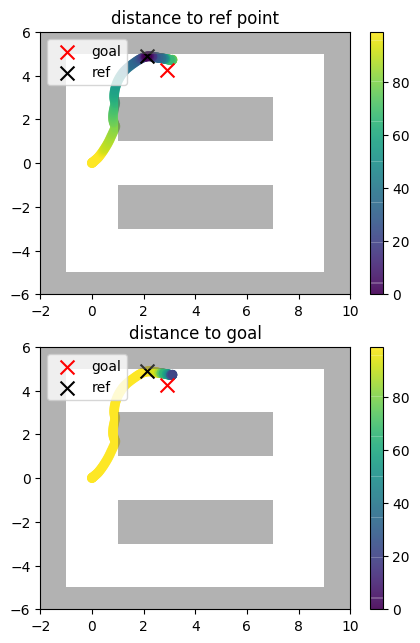}}
    \subfloat[]{\includegraphics[trim={0.5cm 12cm 0 1cm}, clip, width=0.15\textwidth, height=2cm]{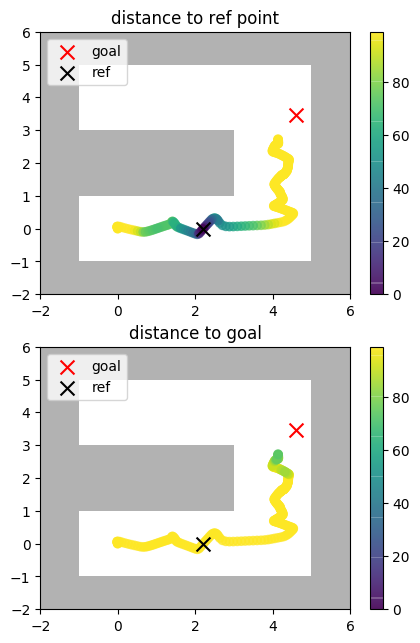}}
    \subfloat[]{\includegraphics[trim={0.5cm 1cm 3cm 20.5cm}, clip, width=0.28\textwidth, height=2cm]{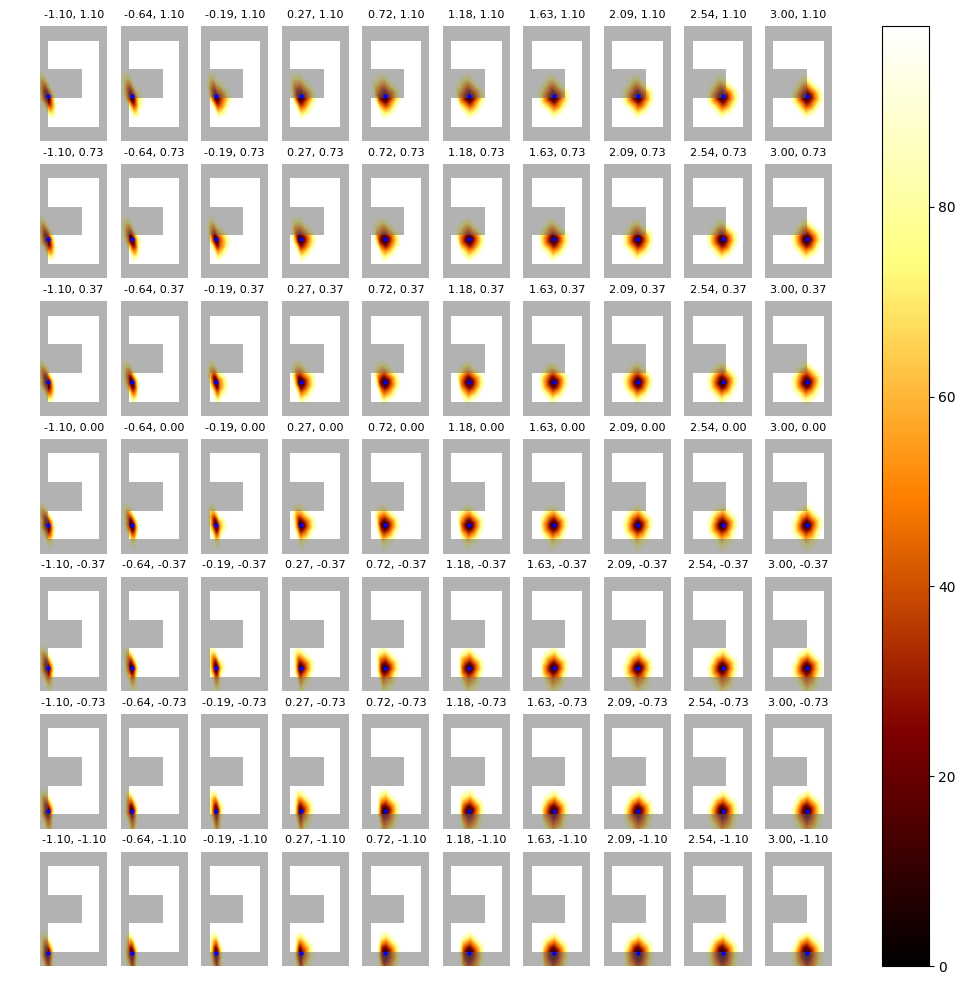}}
    \subfloat[]{\includegraphics[trim={0.5cm 0 4cm 13.5cm}, clip, width=0.28\textwidth, height=2cm]{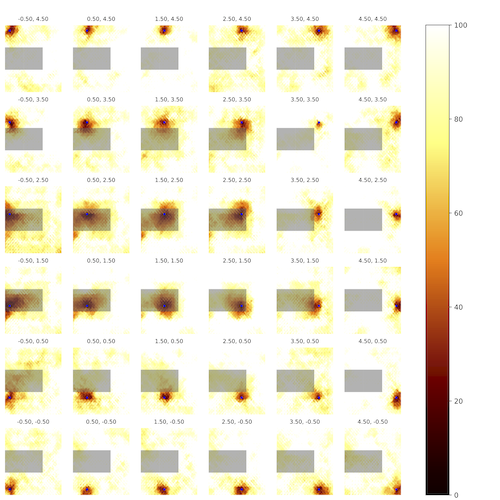}}
   \end{minipage}
    \caption{(a, b): Predicted action distance between a reference state and states along a trajectory in Point Mass (a) and Maze Ant (b); (c,d,e): Predicted action distance between stabilized reference states and all other states. (c): Ant Maze environment in full goal space (trained online) and (d): pixel space (batch setting) respectively. Closer states are darker. Each subplot uses a different reference state near the starting state shown as a blue dot. Full heatmap for (d) is shown in Figure \ref{fig:ours_pixel}.}
    \label{fig:visualization_plots}
\end{figure*}

Results for these experiments are shown in Figures \ref{fig:visualization_xy_goal_space} and \ref{fig:visualization_full_state}, where we can see that the agents are able to make progress on all tasks. For the Point Mass agent, progress is slow compared to the $(x, y)$ case, since now to achieve a goal the agent has to reach a specified position with a specified velocity. The learning progress of the Ant agent trained to reach only stable positions suggests that our approach of learning the distance function is robust to unseen states, since the distance predictor can generalize to states not seen or only rarely seen during training. Fig. \ref{fig:visualization_plots}(a,b) shows the visualization of the distance estimate of our predictor on states visited during a sample trajectory in the Point Mass and Maze Ant environments in $(x, y)$ and full goal spaces respectively. Fig. \ref{fig:visualization_plots}(c) shows the visualization on a set of reference states in the Maze Ant environment in full goal space. We observe that in all experiments, including the set with the $(x, y)$ goal space, the on-policy and the off-policy methods for training the distance predictor performed similarly. In section \ref{sec:expanding_sphere} we study the qualitative difference between the on-policy and off-policy methods for training the distance predictor.


\subsection{Pixel inputs\label{sec:visual_experiment}}
To study whether the proposed method of learning a distance function scales to high-dimensional inputs, we evaluate the performance of the distance predictor using pixel representation of the states. This experiment is performed in the batch setting. Similar to the pretraining phase of \cite{wu2018the}, the embeddings are trained using sample trajectories collected by randomizing the starting states of the agent. Each episode is terminated when the agent reaches an unstable position or $100$ time steps have elapsed. Figure \ref{fig:visualization_plots} (d) shows the distance estimates of the distance predictor in this setting. For qualitative comparison, we also experimented with the approach proposed in \cite{wu2018the}; these results are shown in section \ref{sec:visual_exp_appendix} for various choices of hyperparameters. 

\subsection{Expanding $\epsilon$-sphere} \label{sec:expanding_sphere}
In this section we show that there are qualitative differences between the on-policy and off-policy schemes for training the distance predictor. Since the goal is considered achieved when the agent is within the $\epsilon$-sphere of the goal, the episode is terminated when the agent reaches the boundary of the $\epsilon$-sphere. As the learning progresses and the agent learns a shortest path to the goal, the agent only learns a shortest path to a state on the boundary of the $\epsilon$-sphere of the corresponding goal. In this scenario, the path to the goal $g$ from any state within the $\epsilon$-sphere of $g$ under the policy conditioned on $g$ need not necessarily be optimal since such trajectories are not seen by the policy conditioned on that specific goal $g$. However, the number of actions required to reach the goal $g$ from the states outside the $\epsilon$-sphere along the path to the goal decreases as a result of learning a shorter path due to policy improvement. Therefore, as the learning progresses until an optimal policy is learned, the number of states from which the goal $g$ can be reached in a fixed number of actions increases, thus resulting in an increasing the volume of the $\epsilon$-sphere centered on the goal for a fixed action distance $k$, when using on-policy samples to learn the distance predictor.

\begin{figure}
    \centering
     \includegraphics[width=0.5\textwidth]{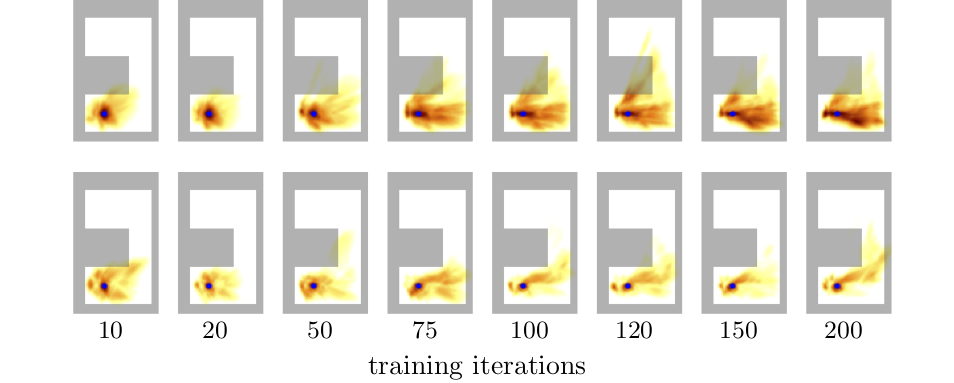}
    \caption{Predictions of the distance predictor trained with on-policy (top) and off-policy (bottom) samples in Maze Ant with $(x, y)$ goal space illustrating how the predictions evolve over time. Darker colors indicate smaller predicted distance and the small blue dot indicates the reference state.}
    \label{fig:expanding_eps_sphere}
\end{figure}

This phenomenon is empirically illustrated in the top row in Fig. \ref{fig:expanding_eps_sphere}. For a fixed state $g$ near the starting position, the distance from all other states to $g$ is plotted. The evolution of the distance function over iterations shows that for any fixed state $s$, $d^\pi(s, g)$ gets smaller. Equivalently, the $\epsilon$-sphere centered on $g$ increases in volume. In contrast, the bottom row in Fig. \ref{fig:expanding_eps_sphere} illustrates the predictions made by an off-policy distance predictor; in that case, the dark region is almost always concentrated densely near $g$, and the volume of the $\epsilon$-sphere exhibits significantly less growth.

Since the agent does not receive training for states that are within the $\epsilon$-sphere centered at a goal $g$, it is desirable to keep the $\epsilon$-sphere as small as possible. One way to do this would be to employ an adaptive algorithm for choosing $\epsilon$ as a function of $g$ and the agent's estimated skill at reaching $g$; as the agent gets better at reaching $g$, $\epsilon$ should be lowered. We leave the design of such an algorithm for future work, and propose the off-policy scheme as a practical alternative in the meantime. We note that this phenomenon is not observed in the visualization in the full goal space, possibly due to the stabilization of the ant during evaluation.


\subsection{Generating Goals Using Action Noise}
\begin{figure}[!tb]
  \begin{minipage}{0.48\textwidth}
    \centering
    \subfloat[Maze Ant]{\includegraphics[width=0.3\linewidth]{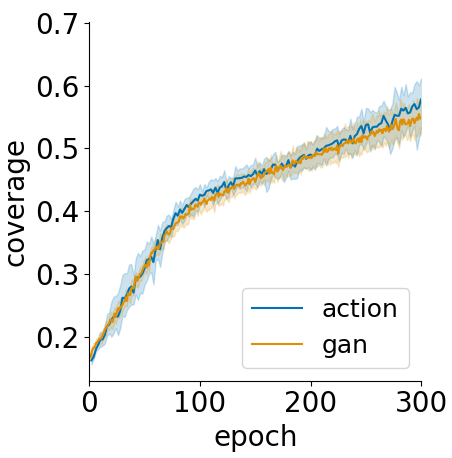}}
    \subfloat[Free Ant]{\includegraphics[width=0.3\linewidth]{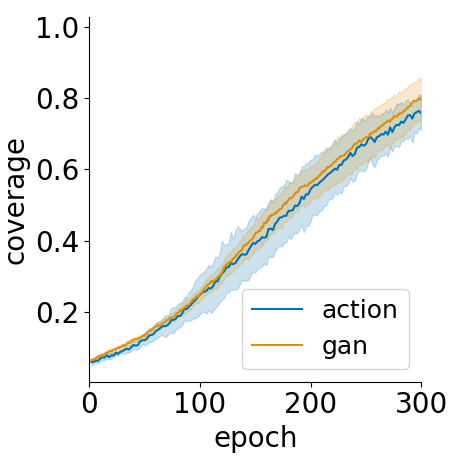}}
    \caption{Comparing the proposed goal generation algorithm against GoalGAN.}
    \label{fig:goal_gen_plot}
  \end{minipage}
\end{figure}

We perform this comparison in both the Maze Ant and Free Ant environments, using $(x, y)$ as the goal space and the Euclidean distance. The results, shown in Fig. \ref{fig:goal_gen_plot}, demonstrate that the performance of our approach is comparable to that of GoalGAN while not requiring the additional complexity introduced by the GAN. The evolution of the working set of goals maintained by our algorithm for Maze Ant is visualized in Fig. \ref{fig:maze_goal_evolution}. 

Though our approach requires additional environment interactions, it does not necessarily have a higher sample complexity compared to GoalGAN in the case of indicator reward functions. This is because the goals generated by GoalGAN are not guaranteed to be feasible (unlike our approach); trajectories generated for unfeasible goals will receive 0 reward and will not contribute to learning.

{}

%% file: icml2020/sections/conclusion.tex
We have presented an approach to automatically learn a task-specific distance function without the requirement of domain knowledge, and demonstrated that our approach is effective in the online setting where the distance function is learned alongside a goal-conditioned policy while also playing a role in training that policy. We then discussed and empirically demonstrated  the expanding $\epsilon$-sphere phenomenon which arises when using the on-policy method for training the distance predictor. This can cause difficulty in setting the $\epsilon$ hyperparameter, particularly when the final performance has to be evaluated using the learned distance function instead of using a proxy evaluation metric like the Euclidean distance. This indicates that off-policy distance predictor training should be preferred in general. Finally, we introduced an action space goal generation scheme which plays well with off-policy distance predictor training. These contributions represent a significant step towards making goal-conditioned policies applicable in a wider variety of environments (e.g. visual domains), and towards automating the design of distance functions that take environment dynamics into account without using domain knowledge.
\clearpage

%% file: icml2020/sections/appendix.tex
\newpage
This is the supplementary for the paper titled "Self-supervised Learning of Distance Functions for Goal-Conditioned Reinforcement Learning"
\appendix
\section{Discussion of MDS and Spectral embedding \label{sec:discussion}}
\input{icml2020/sections/appendices/discussion.tex}

\section{Training the distance predictor}
\input{icml2020/sections/appendices/train_predictor.tex}

\section{Goal Generation Buffer}
\input{icml2020/sections/appendices/goalbuffer.tex}

\section{Hyperparameters}
\input{icml2020/sections/appendices/hyperparams.tex}

\section{Goal Stabilization for Ant Tasks}\label{sec:stable}
For Ant tasks we used a modified setup for obtaining goals from states. We first identified a stable pose for the Ant's body, with body upright and limbs in standard positions. Then, to create a goal from a state, we take the position component from  the state, but take all other components (joint angles and angular velocities) from the stable pose. This stabilization step is used for all goals; however, the distance predictor is still trained on the full state space.

\section{Further experiments \label{sec:additional_experiments}}
\input{icml2020/sections/appendices/furtherexps.tex}


%% file: icml2020/sections/appendices/discussion.tex
\subsection{Classical MDS \label{sec:cmds_appendix}}
First we discuss the tools used in cMDS, before describing cMDS itself. 

Provided with a matrix $X \in \mathbf{R}^{n \times m}$ of $n$ objects in $m$-dimensional space, we can form the squared pairwise distances matrix denoted by ${D^{(2)} \in \mathbf{R}^{n \times n}}$ such that ${D^{(2)}_{ij}=||x_i - x_j||^{2}}$ and can be expressed as $D^{(2)} = c\mathbf{1}^{T} + \mathbf{1}c^{T} - 2XX^{T}$ where $c \in \mathbf{R}^n$ such that $c_i = ||x_i||^{2}$ and $\mathbf{1} \in \mathbf{R}^n$ is the all-ones vector of length $n$.
Let $B = XX^{T}$. The squared pairwise distances matrix $D^{(2)}$ and the scalar product matrix $B$ can be obtained from the configuration matrix $X$. 

Recovering $X$ from $D^{(2)}$ is the objective of cMDS. We first consider the simpler case of recovering $X$ from the scalar product matrix $B$. Since $B$ is symmetric and positive semi-definite, $B$ admits an eigen-decomposition
\begin{equation}
    \label{eq:similarity_factorization}
    B = Q \Lambda Q^{T}=Q\Lambda^{\frac{1}{2}}\Lambda^{\frac{1}{2}}Q^{T} = X^{'}{X^{'}}^{T}
\end{equation}
where $\Lambda_{ii}$ is the $i^{th}$ largest eigenvalue of $B$ and $Q$ is an orthogonal matrix whose columns consist of eigenvectors of $B$ ordered by their corresponding eigenvalues in descending order. Hence, given a pairwise scalar product matrix $B$, a configuration $X^{'}$ that preserves the pairwise distances can be recovered. The origin is assumed to be $\mathbf{0} \in \mathbf{R}^n$.

We now proceed to describe the procedure to obtain a configuration that preserves squared pairwise distances given in $D^{(2)}$.  Let $J=I - \frac{1}{n}\mathbf{1}\mathbf{1}^T \in \mathbf{R}^{n \times n}$. For any $x \in \mathbf{R}^n$, $y=Jx$ is a column vector in $\mathbf{R}^n$ balanced on the origin (mean is zero). Similarly for $x \in \mathbf{R}^{1 \times n}$, $xJ$ is a row vector balanced on the origin. Since $D^{(2)}= c\mathbf{1}^{T} + \mathbf{1}c^{T} - 2XX^{T}$ we get
\begin{align*}
\label{eq:double_center_D}
    -\frac{1}{2}JD^{(2)}J &= -\frac{1}{2}J(c\mathbf{1}^{T} + \mathbf{1}c^{T} - 2XX^{T})J \\
    &= 0 - 0 + JXX^{T}J \numberthis
\end{align*}
because $J\mathbf{1}=0$. Since we are only interested in a configuration $X$ that preserves the distance by the application of cMDS, we assume that the columns of $X$ are balanced on $0$. Therefore $JX=X$. Hence, equation (\ref{eq:double_center_D}) is written as $-\frac{1}{2}JD^{(2)}J=XX^T=B$.  Following equation (\ref{eq:similarity_factorization}), an eigen-decomposition is performed on $B=Q\Lambda^{\frac{1}{2}}\Lambda^{\frac{1}{2}}Q^{T}$. Let $\Lambda^{\frac{1}{2}}_{+}$ be the matrix containing columns corresponding to the positive eigenvalues and $Q_+$ be the corresponding eigenvectors. A configuration that preserves the pairwise distances is then obtained $X=Q_{+}\Lambda^{\frac{1}{2}}_{+}$. Note that the reconstruction produces only a configuration that preserves the pairwise distance exactly and does not necessarily recover the original configuration (differs by rotation).

An alternate characterization of cMDS is given by the loss function $L(X)=||XX^{T} - B||_{F}^{2}$ known as \textit{strain} where $F$ is the Frobenius norm. Due to the Frobenius norm, strain can be written as  
\begin{equation}
L(X)=\sum_{i < j} (x_{i}x_{j}^{T} - B_{ij})^2
\end{equation}
showing that cMDS can be used in an iterative setting.

\subsection{Spectral Embeddings of Graphs}
\label{sec:spectral_graph_drawing}
Given a simple, weighted, undirected and connected graph $G$, the laplacian of the graph is defined as $L=D-W$ where $W$ is the weight matrix and $D$ is the degree matrix. The eigenvectors corresponding to the smallest positive eigenvalues of the graph laplacian are used to obtain an embedding for the nodes and have been shown useful in several applications such as spectral clustering \cite{tutospecclustr} and spectral graph drawing \cite{Koren:2003:SGD:1756869.1756936}. 

The discussion here in the finite state setting but the definition of the Laplacian used here is same as that in \cite{wu2018the} and hence we refer the reader to \cite{wu2018the} for a discussion in continuous state spaces.

First we begin by translating the equations in \cite{wu2018the} to the finite state setting. In \cite{wu2018the}, the matrix $D$ serves the dual purpose of being the weight $W$ matrix for the Laplacian $L=S-W$ where $S$ is the degree matrix $S_{ii}=\sum_j W_{ij}$, and the density of the transition distributions of the random walk on the graph. 

In the finite state case we write $\mathbf{P}$ and $\mathbf{W}$ to denote the use of $D$ for transition matrix and weight matrix respectively. $P^\pi$ denotes the transition probabilities of the markov chain induced by the policy $\pi$. $\hat{P}^\pi$ denotes the transition probabilities of the time-reversed markov chain of policy $\pi$. We assume that the stationary distribution $\rho$ of $P^{\pi}$ exists.

The definition of density $D(u,v)$ in \cite{wu2018the} is given by
\begin{equation}
\label{eq:D_kernel}
    D(u,v) = \frac{1}{2} \frac{P^\pi(v|u)}{\rho (v)} + \frac{1}{2} \frac{P^\pi(u|v)}{\rho (u)} \\ 
\end{equation}
Since $D(u,v)$ is integrated with respect to measure $\rho(v)$ and since integrating with respect to a measure is analogous to weighted sum, equation (\ref{eq:D_kernel}) is multiplied by $\rho(v)$ to obtain
\begin{align}
\label{eq:transition_kernel}
    \mathbf{P}_{uv} &= \frac{1}{2} P^\pi(v|u) + \frac{1}{2} \frac{\rho (v)}{\rho (u)}P^\pi(u|v)\\
    &= \frac{1}{2}P^\pi(v|u) + \frac{1}{2}\hat{P}^\pi(v|u)
\end{align}
For obtaining $\mathbf{W}$, notice that $D(u,v)$ is integrated with respect to $\rho(u)$ and $\rho(v)$ in equation (2) of \cite{wu2018the}. Therefore, by multiplying equation (\ref{eq:D_kernel}) by $\rho(u)$ and $\rho(v)$ we obtain
\begin{align}
\label{eq:laplacian_weight}
    \mathbf{W}_{uv} &= \frac{1}{2}\rho(u)\rho(v) \frac{P^\pi(v|u)}{\rho (v)} + \frac{1}{2} \rho(u)\rho(v) \frac{P^\pi(u|v)}{\rho (u)}\\
    &= \frac{1}{2}\rho(u) P^\pi(v|u) + \frac{1}{2} \rho(v) P^\pi(u|v)
\end{align}
Note that the transition probabilities $\mathbf{P}$ is given by forward transition probabilities $P^{\pi}$ with $0.5$ probability and the time-reversed $\hat{P}^{\pi}$ with $0.5$ probability.  Such a restriction is required since transition probabilities of a random walk on an undirected graph have to be reversible. However, in the RL settting the samples are collected only using $P^{\pi}$. Hence, we assume that $P^{\pi}$ is reversible. Note that this assumption is a special case of the transition matrix $\mathbf{P}$ given in equation (\ref{eq:transition_kernel}). Hence, the definition of $\mathbf{P}_{uv}$ is given by $\mathbf{P}_{uv}=P^{\pi}(u|v)$ and $\mathbf{W}$ is given by $\mathbf{W_{uv}}=\rho(u)P^{\pi}(v|u)$.

\subsection{Euclidean Commute Time Distance\label{sec:ectd}}
Given the Laplacian $L=D-W$, \cite{Fouss:2005:NWC:1092358.1092536, 10.1109/TKDE.2007.46} show that the average first passage time and the average commute times can be expressed as
\begin{equation}
    \label{eq:avg_first_pass_laplacian_inv}
    m(j|i) = \sum_{k=1}^{n} (l_{ik}^{\dagger} - l_{ij}^{\dagger} - l_{jk}^{\dagger} + l_{jj}^{\dagger}) d_{kk}
\end{equation}
and 
\begin{equation}
    \label{eq:avg_commpute_laplacian_inv}
    n(i, j) = V_{G}(l_{ii}^{\dagger} + l_{jj}^{\dagger} - 2 l_{ij}^{\dagger})
\end{equation}
respectively, where $V_G = \sum_{i=1}^{n} D_{ii}$ is the volume of the graph.
$n(i, j)$ can be expressed as 
\begin{align*}
    n(i, j) &= V_G(l_{ii}^{\dagger} + l_{jj}^{\dagger} -  l_{ij}^{\dagger} -  l_{ij}^{\dagger}) \\
    &= V_G (e_i - e_j)^T L^{\dagger} (e_i - e_j)\\
    &= V_G (e_i - e_j)^T  Q \Lambda Q^T (e_i - e_j) \\
    &= V_G (e_i - e_j)^T  Q \Lambda^{\frac{1}{2}} \Lambda^{\frac{1}{2}} Q^T (e_i - e_j) \\ 
    &=V_G (e_i - e_j)^T  Q \Lambda^{\frac{1}{2}T} \Lambda^{\frac{1}{2}} Q^T (e_i - e_j)\\
    &=V_G (x_i - x_j)^T (x_i - x_j) \numberthis \label{eq:ectd}
\end{align*}
where $x_k$ is $\Lambda^{\frac{1}{2}}Q^T e_k=(e_k^T Q \Lambda^{\frac{1}{2}})^T$, the column vector corresponding to the $i^{th}$ row of $Q \Lambda^{\frac{1}{2}}$. The $L2$ distance in this embedding space between nodes $(i,j)$, $||x_i - x_j||$, corresponds to $\sqrt{n(i,j)}$ up to a constant factor $\frac{1}{\sqrt{V_G}}$, termed euclidean commute time distance (ECTD) in \cite{Fouss:2005:NWC:1092358.1092536}.

Since the orthogonal complement of the null-space of the linear transformation is invertible and a vector of all ones $\mathbf{1}$ spans the nullspace of $L$, psedudo-inverse of the Laplacian is intuitively given by $L^{\dagger}=(L + \frac{1}{n} \mathbf{1}\mathbf{1}^T)^{-1} - \frac{1}{n} \mathbf{1}\mathbf{1}^T$. Hence, it is easy to verify that $L^{\dagger}$ is symmetric and the nullspace of $L^{\dagger}$ is also $\mathbf{1}$. Therefore, $L^{\dagger}$ is double centered. The resultant matrix of the double centering operation $JAJ$ on any matrix $A$ are given by 
\begin{align*}
&(JAJ)_{ij}\\
&= a_{ij} - \frac{1}{n}\sum_{k=1}^{n}a_{ik} - \frac{1}{n}\sum_{m=1}^{n}a_{mj} + \frac{1}{n^2}\sum_{m=1}^n\sum_{k=1}^{n}a_{mk}
\end{align*}
Using these two facts, it is easy to verify that 
\begin{align*}
    &-\frac{1}{2}(JNJ)_{ij}\\
    &= L^{\dagger}_{ij} - \frac{1}{n}\sum_{k=1}^{n}L^{\dagger}_{ik} - \frac{1}{n^2}\sum_{k=1}^{n}L^{\dagger}_{kk} + \frac{1}{n^2}\sum_{m=1}^{n}\sum_{k=1}^{n}L^{\dagger}_{mk}\\
    &=L^{\dagger}_{ij} \quad \text{(since $L^{\dagger}$ is double centered)}
\end{align*}
where $N$ is the matrix of commute times among all pairs of nodes.

Hence, the solution to the embedding space that preserves ECTD given by equation (\ref{eq:ectd}) is the same as the one provided by classical MDS by taking $N$ as $D^{(2)}$ or $L^{\dagger}$ as $B$. 

Finally, since that the eigenvectors of $L$ and $L^{\dagger}$ are the same and the non-zero eigenvalues of $L^{\dagger}$ is the inverse of the corresponding non-zero eigenvalues of $L$. This shows that simply using the eigenvectors of the Laplacian is insufficient to obtain an embedding where the distances are preserved. An empirical comparision of the embeddings obtained from eigenvectors and the scaled eigenvectors of the Laplacian to compute distances is shown in \ref{sec:compare_laplacian_pseudo_laplacian} in a tabular setting.

\subsection{Approximation using Metric MDS \label{sec:approx_with_mmds}}

The mean first passage times $m(.|.)$ are not available and has to be estimated from the trajectories. First note that $m(j|i)=\sum_{t=0}^{\infty}P_{ij}^t t$ where we use the notation $P_{ij}^t$ to denote the probability of going from state $i$ to $j$ in exactly $n$ steps for the first time. As a result, the objective function is of the form 
\begin{equation}
    \sigma(X)=\frac{1}{2}\sum_{i=1}^n\sum_{j=1}^n w_{ij} (||x_i - x_j||_2^2 - k)^{2}
    \label{eq:stress_m_estimate}
\end{equation}
where $k \sim P_{ij}^t$.
The quantities in equation (\ref{eq:stress_m_estimate}) can be obtained from trajectories drawn under a fixed policy, thus providing a practical approach to learn the embedding as given in equation (\ref{eq:1}) with $q=2$. When the quantities $k$ are obtained from the trajectories, in addition to the weights $w_{ij}$ on $(i,j)$, each $k$ is weighted by $P_{ij}^k$. This emphasizes the shorter distances for each pair of $(i,j)$. As show in \ref{sec:q_effect}, $q$ in equation (\ref{eq:1}) provides a mechanism to control the trade-off between the larger and smaller distances by considering $q$ as a hyperparameter. Thus, setting $w_{ij}^k$ to $\rho(i) P_{ij}^k + \rho(j) P_{ji}^k$ provides a practically convenient set of weights and the trade-off it induces can be mitigated as desired by changing $q$.


%% file: icml2020/sections/appendices/train_predictor.tex
The distance predictor has to be trained prior to being used in determining whether the goal has been reached in order to produce meaningful estimates. To produce a initial set of samples to train the distance predictor, we use a randomly initialized policy to generate a set of samples. This provides a meaningful initialization for the distance predictor since these are states that are most likely to occur under the initial policy. 

The distance predictor is a MLP with $1$ hidden layer and $64$ hidden units and ReLU activation. We initialize distance predictor by training on $100,000$ samples collected according to the randomly initialized policy  for $50$ epochs. The starting position is not randomized. In subsequent iterations of our training procedure the MLP is trained for one epoch. The learning rate is set to $1e$-$5$ and mini-batch size is $32$. The distance predictor is trained after every policy optimization step using either either off-policy or the on-policy samples. The embeddings are $20$-dimensional and we use $1$-norm distance between embeddings as the predicted distance.

%% file: icml2020/sections/appendices/goalbuffer.tex
To generate goals according to the proposed approach we store the states visited under the random policy after reaching the goals in a specialized buffer and sample uniformly from the buffer to generate the goals for each iteration. The simplest approach of storing the goals in a list suffers from the two following issues: i) all the states visited under the random policy from the beginning of the training procedure will be considered as a potential goal for each iteration and ii) the goals will be sampled according to the state visitation distribution under the random policy. The issues i) and ii) are problematic because the goal generation procedure has to adapt to the current capacity of the agent and avoid the goals that have already been mastered by the agent; sampling goals according to the visitation distribution will bias the agent towards the states that are more likely under the random policy. To overcome these issues we use a fixed size queue and by ensure that the goals in the buffer are unique. To avoid replacing the entire queue after each iteration, only a fixed fraction of the states in the queue are replaced in each iteration. In our experiments, the queue size was set $500$ and $30$ goals were updated after each iteration.

We observe that our goal generation or off-policy distance predictor are agnostic to the policy being a random policy and hence the random policy can be replaced with any policy if desired.

\begin{figure}[!htb]
    \begin{minipage}{0.48\textwidth}
    \captionsetup[subfigure]{labelformat=empty}
    \centering
    \includegraphics[width=0.98\textwidth]{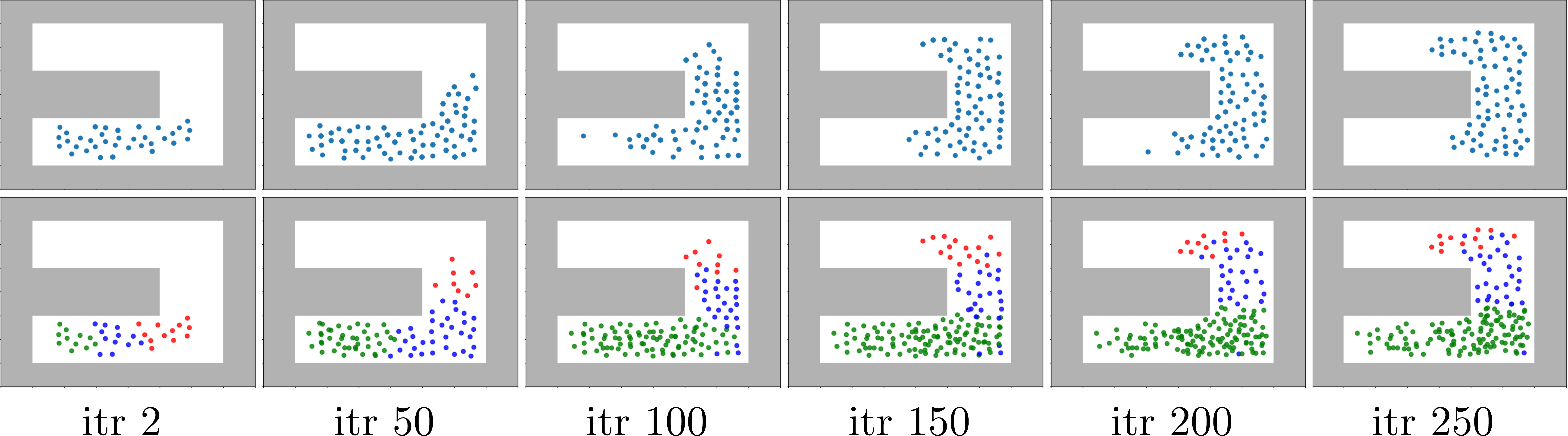}
    \caption{Evolution of the goals generated by our goal generation approach (top). A sample of goals so-far encountered (bottom),
color-coded according to estimated difficulty: green are easy, blue are GOID and red are hard.}
    \label{fig:maze_goal_evolution}
    \end{minipage}
\end{figure}

%% file: icml2020/sections/appendices/hyperparams.tex
The GoalGAN architecture and its training procedure and the policy optimization procedure in our experiments are similar to \cite{Florensa2018AutomaticGG}. Similar to the distance predictor, GoalGAN is trained initially with samples generated by a random policy. The GAN generator and discriminator have $2$ hidden layers with $264$ and $128$ units respectively with ReLU non-linearity and the GAN is trained for 200 iterations after every $5$ policy optimization iterations. A component-wise gaussian noise of mean zero and variance $0.5$ are added to the output of the GAN similar to \cite{Florensa2018AutomaticGG}. The policy network has $2$ hidden layers with $64$ hidden units in each layer and \textit{tanh} non-linearity and is optimized using TRPO \cite{TRPO} with a discount factor of $0.99$ and \textit{GAE} of $1$. The $\epsilon$ value was set to $1$ and $60$ with $L2$ and the learned distance, respectively, in the Ant environments and $0.3$ and $20$ for $L2$ and learned distance, respectively, for the Point Mass environments. In order to determine the first occurrence of a state, we use a threshold of $1e-4$ in the state space. This value is not tuned.

The hyperparameter for $\epsilon$ and the learning rate were determined by performing grid search with $\epsilon$ values $50, 60$ and $80$ (Maze Ant) and $20, 30$ (Point Mass) and the learning rates of $1e$-$3$, $1e$-$4,1e$-$5$ in the off policy setting. For the sake of simplicity we use the same $\epsilon$ for the on-policy and off-policy distance predictors in our experiments. All the plots show the mean and the confidence interval of $95$\% for all our experiments using $5$ random seeds. Our implementation is based on the github repository for \cite{Florensa2018AutomaticGG}, located at \url{https://github.com/florensacc/rllab-curriculum}.

%% file: icml2020/sections/appendices/furtherexps.tex
\subsection{Effect of scaling the eigenvectors of the Laplacian \label{sec:compare_laplacian_pseudo_laplacian}}
\begin{figure*}[!htbp]
    \centering
    \begin{minipage}{\textwidth}
    \centering
  \subfloat[1]{\includegraphics[width=5.25cm,height=2.cm]{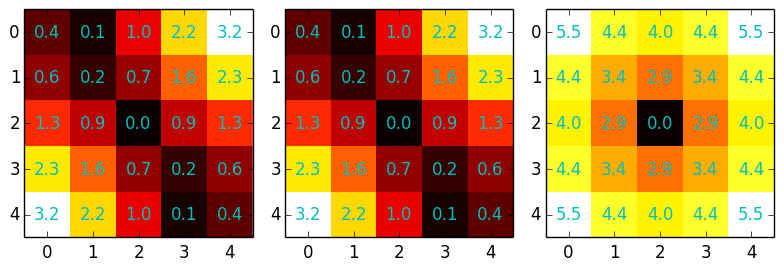}{$\quad$}}
    \subfloat[3]{\includegraphics[width=5.25cm,height=2.cm]{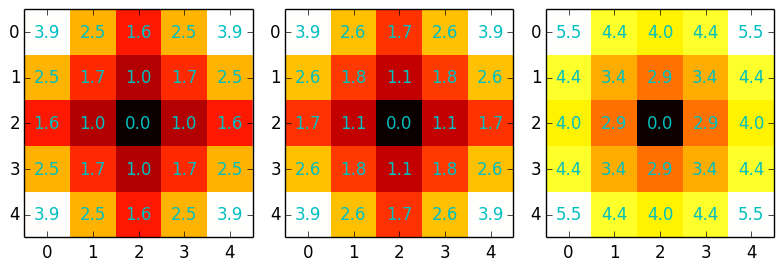}}
  \end{minipage}
\begin{minipage}{\textwidth}
    \centering
  \subfloat[5]{\includegraphics[width=5.25cm,height=2.cm]{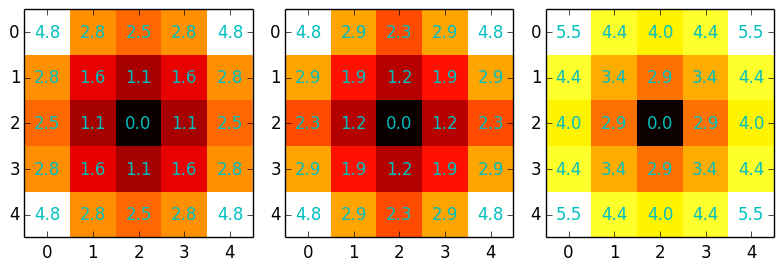}{$\quad$}}
    \subfloat[7]{\includegraphics[width=5.25cm,height=2.cm]{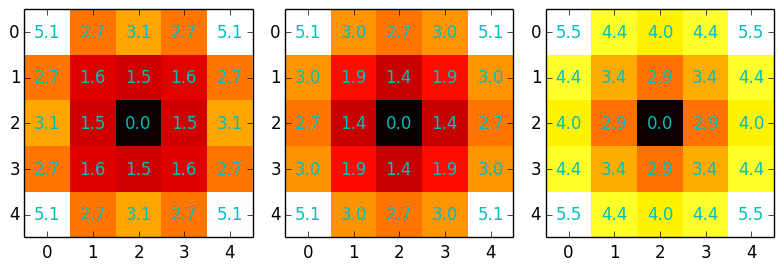}}
  \end{minipage}
\begin{minipage}{\textwidth}
    \centering
  \subfloat[9]{\includegraphics[width=5.25cm,height=2.cm]{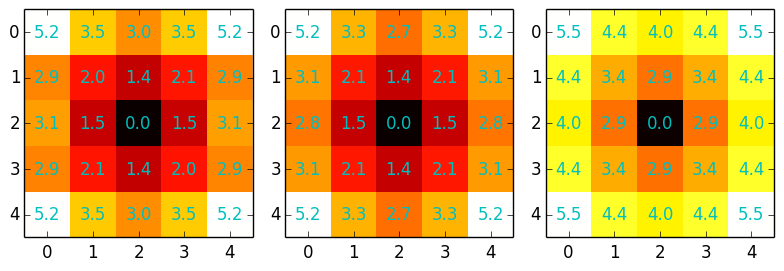}{$\quad$}}
    \subfloat[11]{\includegraphics[width=5.25cm,height=2.cm]{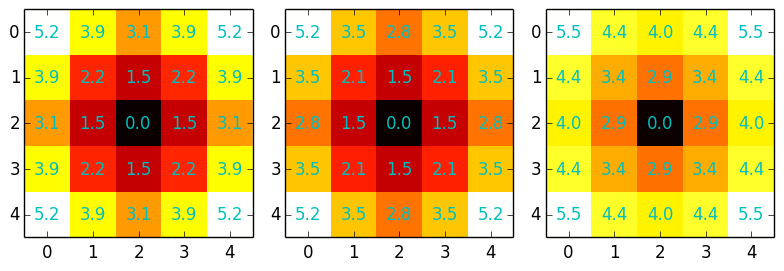}}
  \end{minipage}
\begin{minipage}{\textwidth}
    \centering
  \subfloat[13]{\includegraphics[width=5.25cm,height=2.cm]{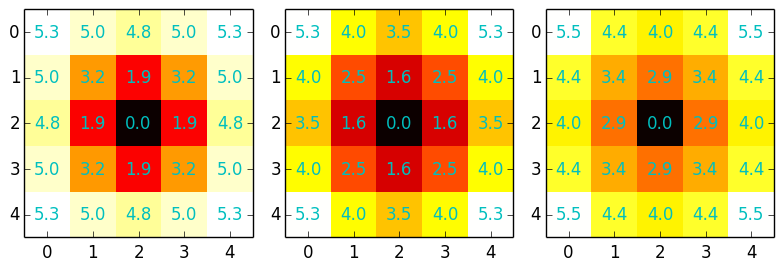}{$\quad$}}
    \subfloat[15]{\includegraphics[width=5.25cm,height=2.cm]{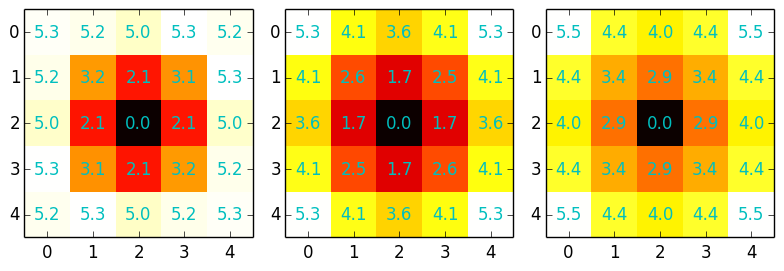}}
  \end{minipage}
\begin{minipage}{\textwidth}
    \centering
  \subfloat[17]{\includegraphics[width=5.25cm,height=2.cm]{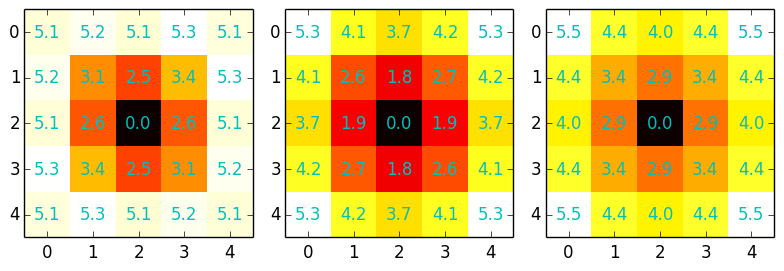}{$\quad$}}
    \subfloat[19]{\includegraphics[width=5.25cm,height=2.cm]{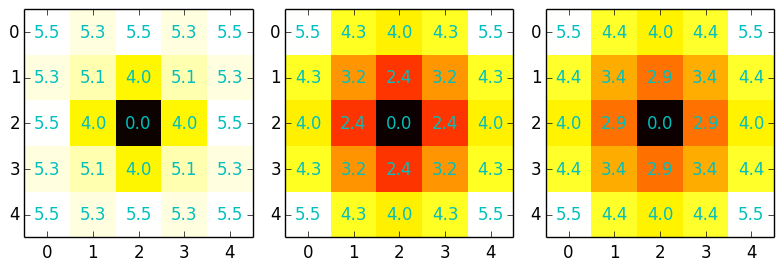}}
  \end{minipage}
\begin{minipage}{\textwidth}
    \centering
  \subfloat[21]{\includegraphics[width=5.25cm,height=2.cm]{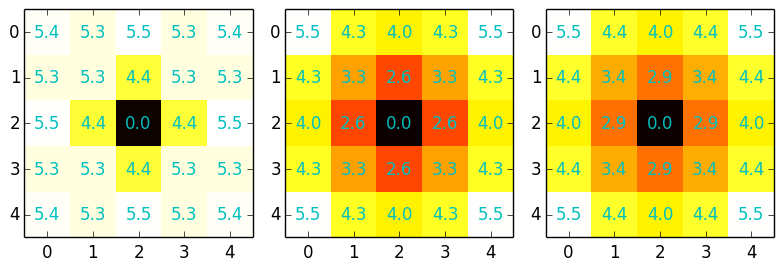}{$\quad$}}
    \subfloat[24]{\includegraphics[width=5.25cm,height=2.cm]{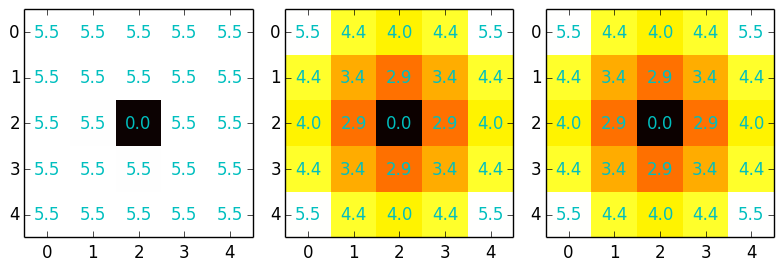}}
  \end{minipage}
\caption{Visualization of the distance from state $(2,2)$ to all other states using different embedding sizes produced by spectral embedding(first column) and scaled spectral embedding(second column) along with the ground-truth computed analytically(third column). The transition from each state to its neighbours are uniformly random. Spectral embedding and scaled spectral embeddings are reasonably similar upto $11$ dimensions. The distance estimate of spectral embeddings deteriorates as many 'less informative' dimensions corresponding to large eigenvalues are added and weighed with as much importance as the 'more informative' dimensions given by small eigenvalues.}
\label{fig:uniform_maze}
\end{figure*}

\begin{figure*}[!htbp]
    \centering
    \begin{minipage}{\textwidth}
    \centering
  \subfloat[1]{\includegraphics[width=5.25cm,height=2.cm]{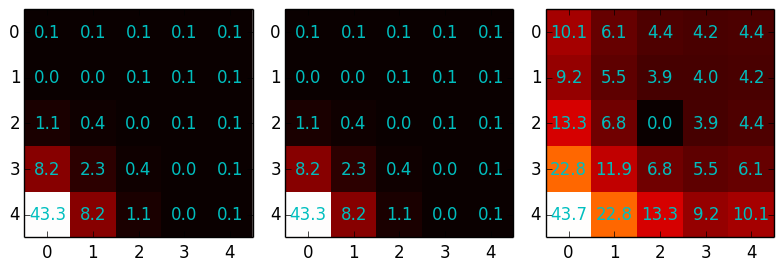}{$\quad$}}
    \subfloat[3]{\includegraphics[width=5.25cm,height=2.cm]{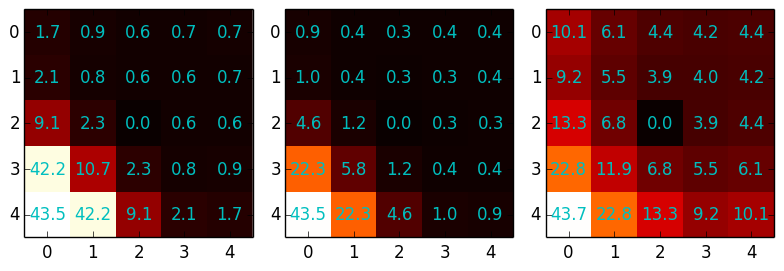}}
  \end{minipage}
\begin{minipage}{\textwidth}
    \centering
  \subfloat[5]{\includegraphics[width=5.25cm,height=2.cm]{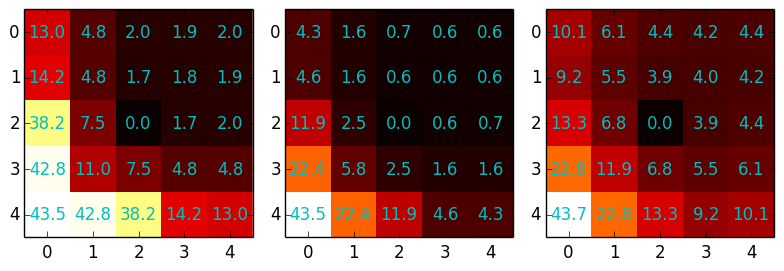}{$\quad$}}
    \subfloat[7]{\includegraphics[width=5.25cm,height=2.cm]{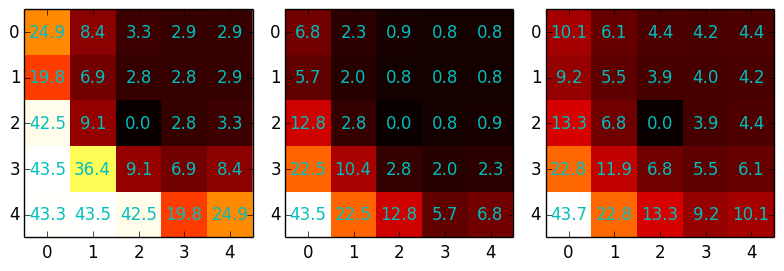}}
  \end{minipage}
\begin{minipage}{\textwidth}
    \centering
  \subfloat[9]{\includegraphics[width=5.25cm,height=2.cm]{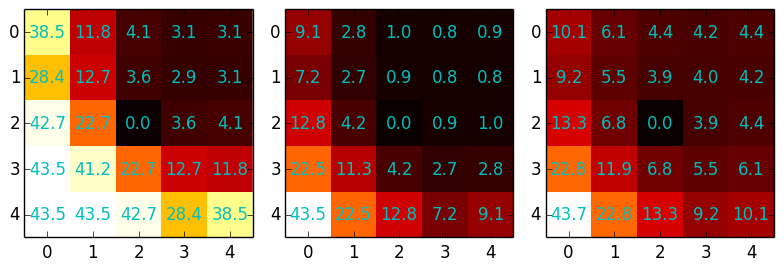}{$\quad$}}
    \subfloat[11]{\includegraphics[width=5.25cm,height=2.cm]{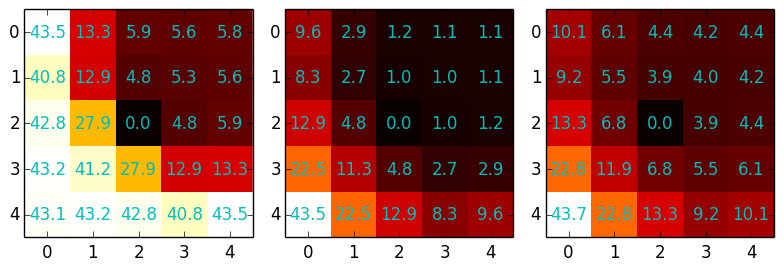}}
  \end{minipage}
\begin{minipage}{\textwidth}
    \centering
  \subfloat[13]{\includegraphics[width=5.25cm,height=2.cm]{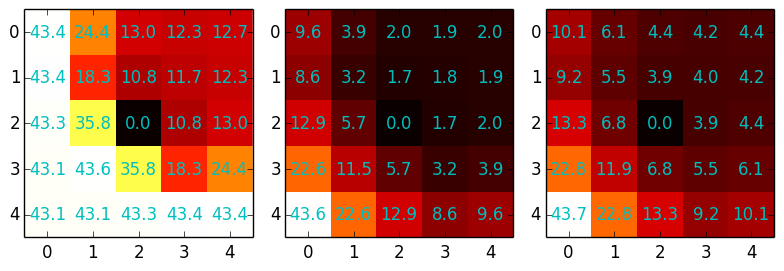}{$\quad$}}
    \subfloat[15]{\includegraphics[width=5.25cm,height=2.cm]{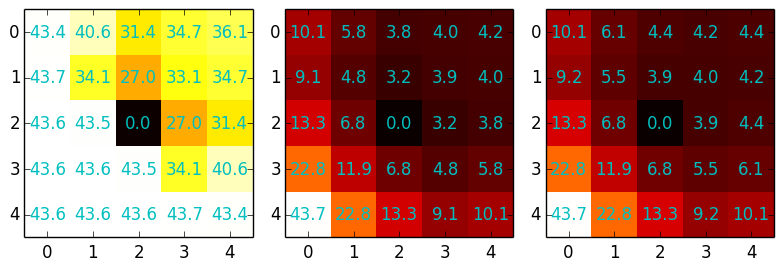}}
  \end{minipage}
\begin{minipage}{\textwidth}
    \centering
  \subfloat[17]{\includegraphics[width=5.25cm,height=2.cm]{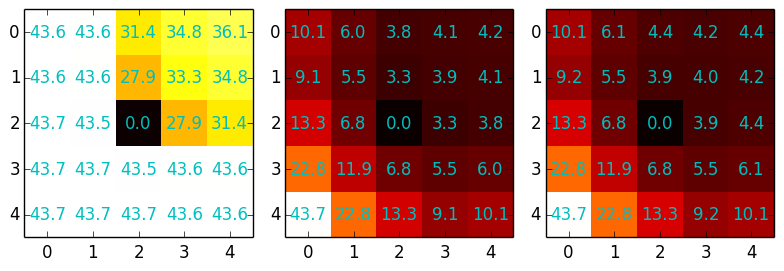}{$\quad$}}
    \subfloat[19]{\includegraphics[width=5.25cm,height=2.cm]{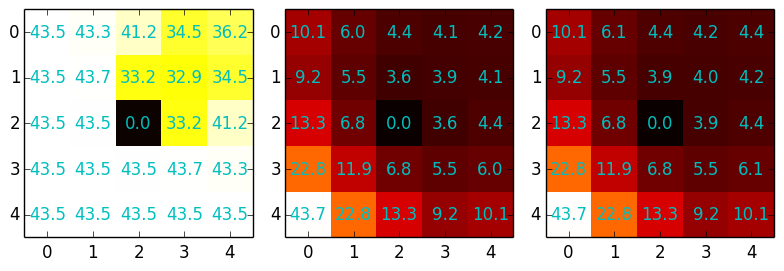}}
  \end{minipage}
\begin{minipage}{\textwidth}
    \centering
  \subfloat[21]{\includegraphics[width=5.25cm,height=2.cm]{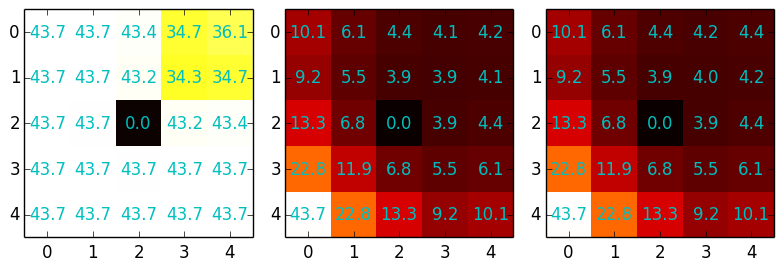}{$\quad$}}
    \subfloat[24]{\includegraphics[width=5.25cm,height=2.cm]{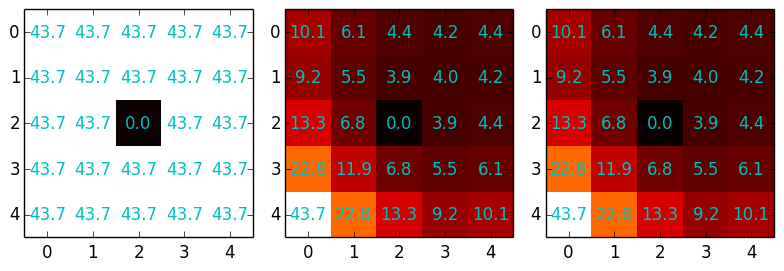}}
  \end{minipage}
\caption{The transition from a state to its neighbours in north and east are with probability $0.375$ and in south and west are with probability $0.125$. The choice of addition of a dimension to spectral embedding presents a trade-off between preserving the previous estimate and improving the estimate of a closer state. In contrast, the scaled spectral embedding improves with the addition of every dimension.}
    
\label{fig:non_uniform_maze}
\end{figure*}

In this section we compare the embeddings obtained using the eigenvectors of the Laplacian (spectral embedding) and the embeddings obtained by scaling the eigenvectors by the inverse of square root of the corresponding eigenvalues (scaled spectral embedding). We perform this comparison in two mazes with $25$ states. A transition from each node to the $4$ neighbours in the north, south, east and west directions are permitted, with equal probabilities in the first maze (figure \ref{fig:uniform_maze}) and, north and east with $0.375$ and south and west with $0.125$ probabilities in the second maze (figure \ref{fig:non_uniform_maze}). We choose the state corresponding to $(2,2)$ as the center and the distance from this state to all the other states are plotted. The first two columns in figures \ref{fig:uniform_maze} and \ref{fig:non_uniform_maze} correspond to spectral embedding and scaled spectral embedding respectively. The third column corresponds to the ground truth $\sqrt{n(i,j)}$ computed analytically from the mean first passage times computed as $M_{ij}=\frac{Z_{jj} - Z_{ij}}{\rho(j)}$ where $Z=(I - \mathbf{P} + \mathbf{1}\rho^T)^{-1}$. The distances produced by spectral embedding are multiplied by the ratio of maximum distance of scaled spectral embedding and the maximum distance of spectral embedding to produce a similar scale for visualization. The Laplacian is given by $L=D-W$ with $W$ given by equation (\ref{eq:laplacian_weight}) and $D$ is the diagonal matrix with stationary probabilities on the diagonal.

\begin{figure}
    \centering
    \subfloat[]{\includegraphics[width=.25\textwidth]{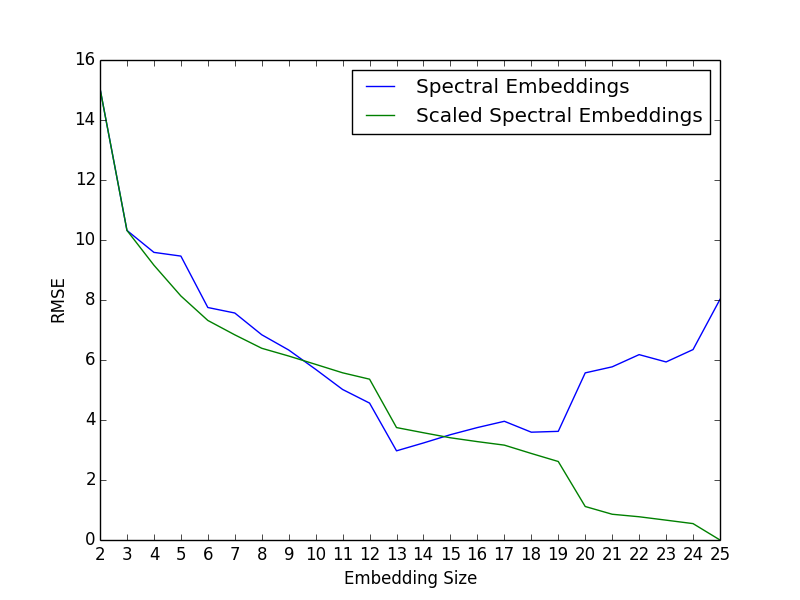}}
    \subfloat[]{\includegraphics[width=.25\textwidth]{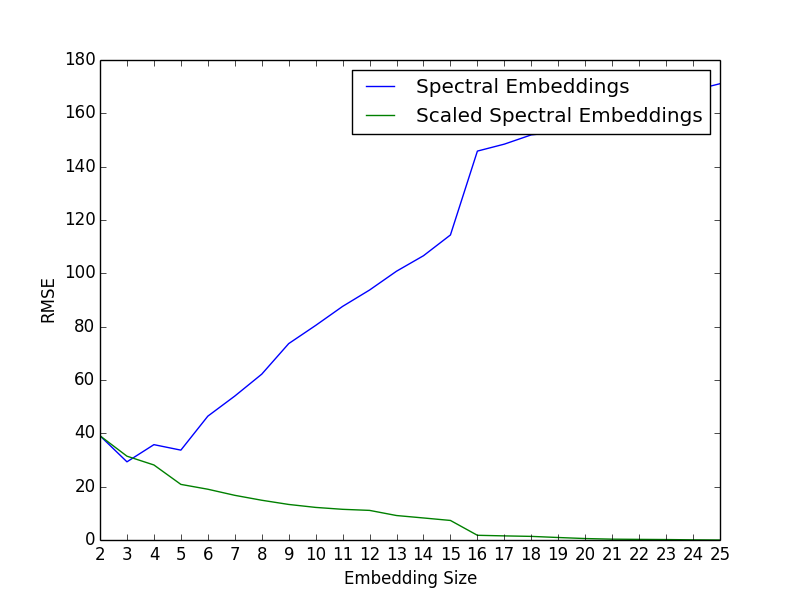}}
    \caption{The square root of commute times from the center state $(2,2)$ to all other states is taken as the reference. RMSE of distances obtained from spectral and scaled spectral embeddings is plotted for (a) Maze with uniform transition probabilities and (b) Maze with non-uniform transition probabilities. The distances obtained from spectral embeddings are scaled by the ratio of maximum distance from scaled spectral embeddings and that of spectral embeddings; this is done to map the spectral embeddings to the same scale as square root of commute times.}
    \label{fig:rmse_comparison}
\end{figure}

As seen in both the Figures \ref{fig:uniform_maze} and \ref{fig:non_uniform_maze}, increasing the size of the embedding dimensions of scaled spectral embeddings better approximates the commute-time distance. The same cannot be said for the approximation given by spectral embeddings. In the first maze (Figure \ref{fig:uniform_maze}), the approximation gets better until the embedding size of $13$ and then deteriorates. When the objective is to find an lower-dimensional approximation of the state space, the choice of the embedding size is treated a hyperparameter and hence one might be tempted to consider this as hyperparameter tuning. However, as shown in the second maze (Figure \ref{fig:non_uniform_maze}), when the environment dynamics are not symmetric, the effect is pronounced to the extent that there is no single choice of the embedding size for the spectral embeddings that best preserves the distances of the nearby states and the faraway states. Even in the case when the embedding size is $3$, the spectral embeddings are markedly different from scaled spectral embeddings as the states $(0,3), (0,4), (1,4)$ are marked equidistant from the reference state by the spectral embeddings. A comparison of the RMSE error of spectral embeddings and scaled spectral embedding is provided in Figure \ref{fig:rmse_comparison}. The difference between spectral embeddings and scaled spectral embeddings is blurred in the uniform transitions case since the eigenvalues are similar. In the non-uniform transitions case, the difference between spectral embeddings and scaled spectral embeddings are evident since the eigenvalues are very dissimilar. Note that similar eigenvalues means the corresponding dimensions have similar weights; scaling by a (approximately) constant - scaled spectral embedding approach - doesn't cause significant difference.

We finally note that our objective is not find a low-dimensional embedding of the state space but to find an embedding that produces a meaningful distance estimate. Scaled spectral embeddings are appropriate for this purpose since the accuracy of the distance estimates improves monotonically with an increase in the number of dimensions.

{}

\subsection{Effect of $q$ \label{sec:q_effect}}
We empirically demonstrate that increasing $q$ increases the effect of larger distances. In order to obtain the same scale of measurements, the distance in the embedding space are raised to the power $q$. We show the effect of $q$ for values $0.5, 1, 2, 4$. It is clearly evident that increasing $q$ increases the radius and the granularity of distance between points that are near and far are lost. The reason for is suggested in \cite{modMDS} (section 11.3). Increasing $q$ increases the weight given to larger distances. For instance, when $q=2$, the stress is approximately $4\delta_{ij}^2(\delta_{ij} - d_{ij})^2$ where $\delta_{ij}=\sqrt{n(i, j)}^{\frac{1}{2}}$ since the target dissimilarity can be rewritten as ${\delta_{ij}^{\frac{1}{q}}}^q$. The $q^{th}$ root of the $\delta_{ij}$ decreases the granularity of the difference between near and far states and the larger weighting term results in overweighting sporadic examples causing an increase in radius. Similarly, it can be shown that when $q=4$, the stress is given by $16\delta_{ij}^6(\delta_{ij} - d_{ij})^2$ where $\delta_{ij}=\sqrt{n(i,j)}^{\frac{1}{4}}$.

\begin{figure*}[hp!]
    \centering
    \subfloat[$q=0.5$]{\includegraphics[trim={0 0 3.9cm 0}, clip, width=7.8cm,height=8.cm]{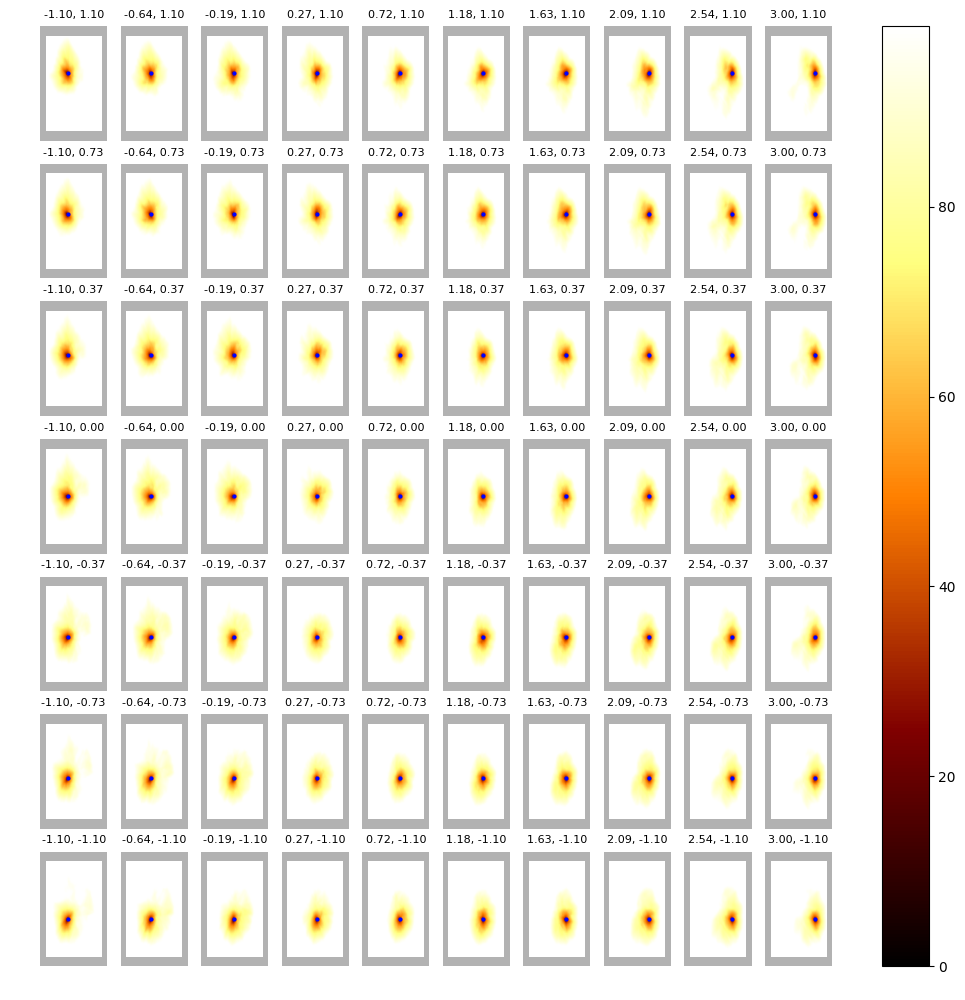}}
    \subfloat[$q=1$]{\includegraphics[width=8.15cm,height=8.cm]{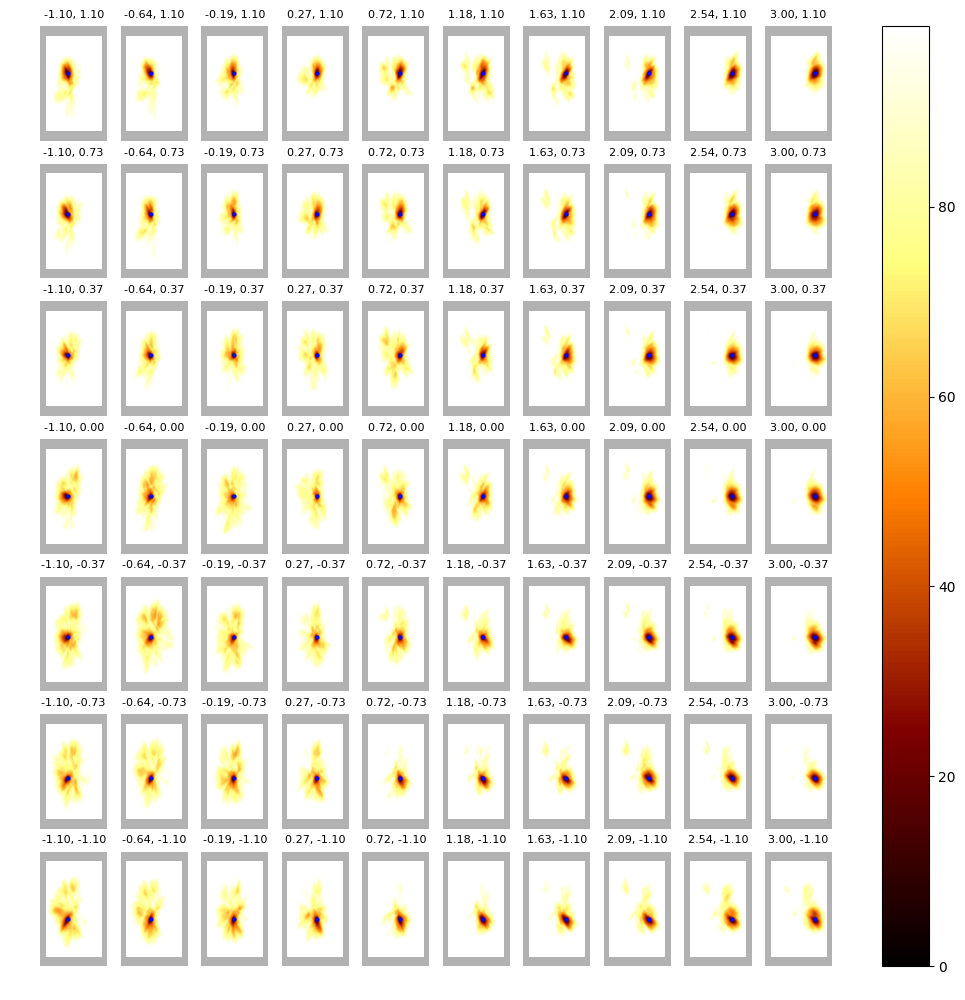}}\\
    \subfloat[$q=2$]{\includegraphics[trim={0 0 3.9cm 0}, clip, width=7.8cm,height=8.cm]{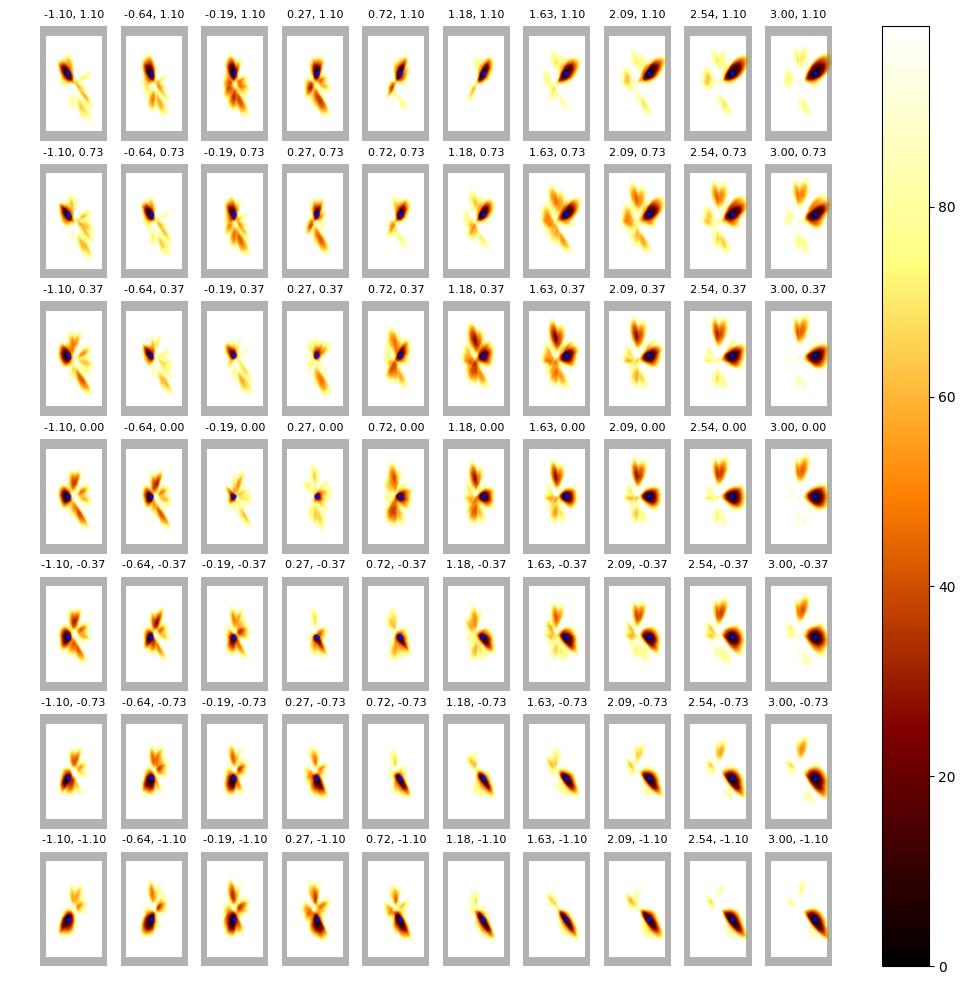}}
    \subfloat[$q=4$]{\includegraphics[width=8.15cm,height=8.cm]{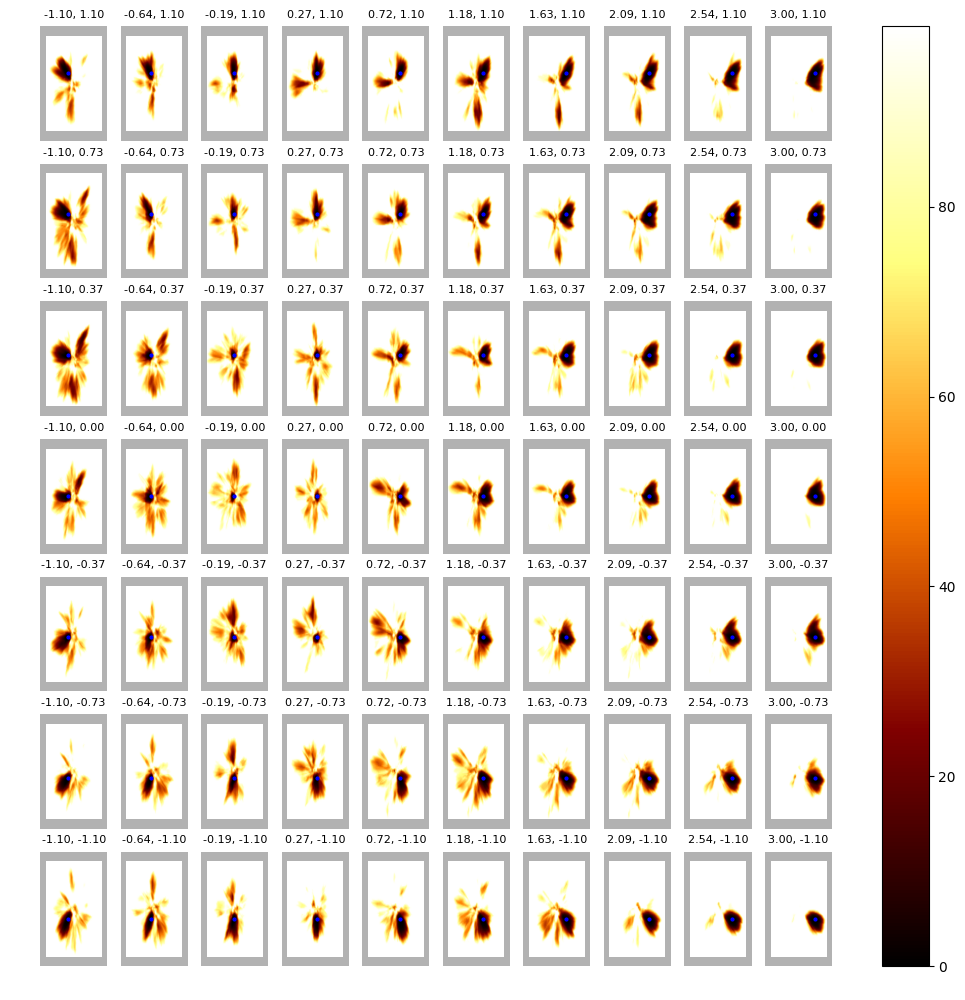}}
    \caption{We study the effect of $q$ on the radius and degree of closeness of the ant agent in the free maze. This shows that $q$ is easy to tune and provides a straightforward mechanism to scale the distances. A similar effect is also observed in the other environments and with pixel inputs.}
    \label{fig:q_effect}
\end{figure*}

\subsection{Prediction with pixel inputs \label{sec:visual_exp_appendix}}

\begin{figure*}
\centering
    \includegraphics[width=5cm,height=5cm]{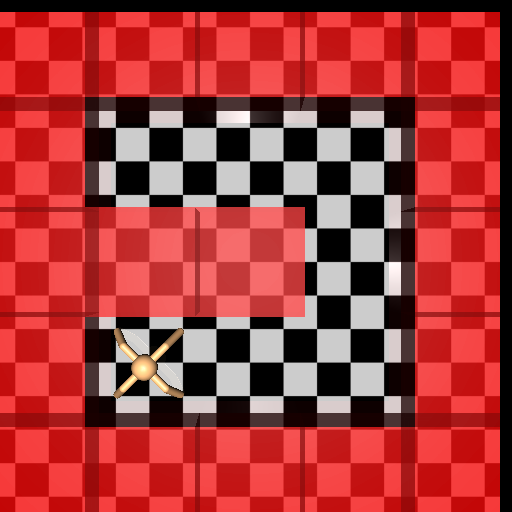}
    \captionof{figure}{Top-down view of the Maze Ant. The RGB image scaled to $32 \times 32$ is the input in the pixel tasks.}
    \label{fig:pixel_top_down_view}
\end{figure*}

\begin{figure*}
    \centering
    \subfloat[$\beta=0.5$]{\includegraphics[trim={0 0 3.9cm 0}, clip, width=8.4cm,height=8.8cm]{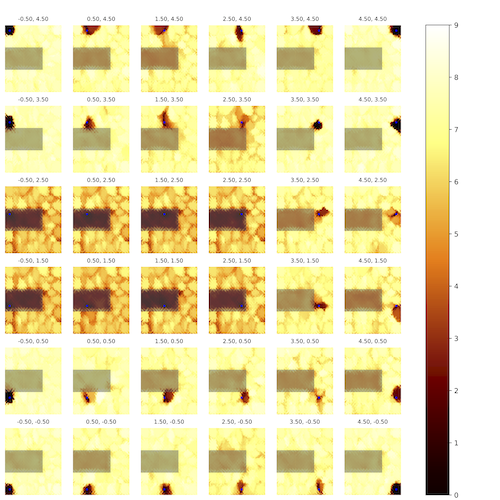}}
    \subfloat[$\beta=1$]{\includegraphics[width=8.6cm,height=8.8cm]{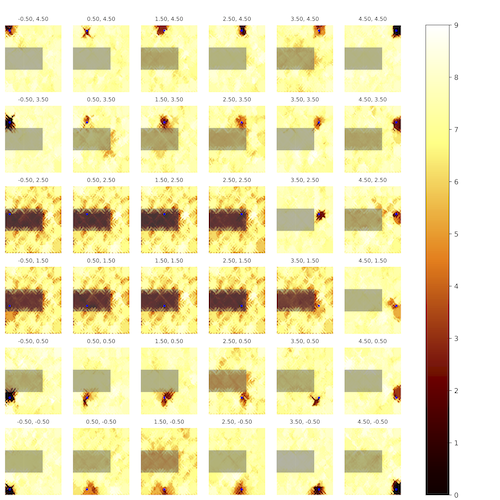}} \\
    \subfloat[$\beta=2$]{\includegraphics[trim={0 0 3.9cm 0}, clip, width=8.4cm,height=8.8cm]{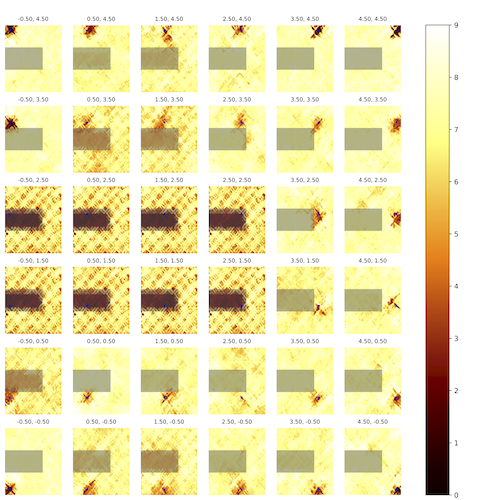}}
    \subfloat[$\beta=6$]{\includegraphics[width=8.6cm,height=8.8cm]{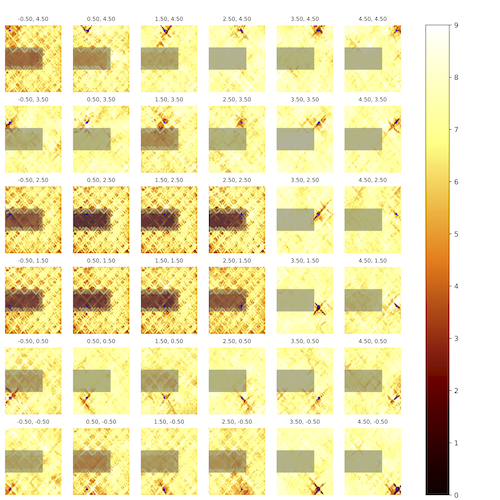}}
    \caption{Laplacian in RL using pixel inputs. Higher $\beta$ better approximates spectral graph drawing objective. $LR=1e-4$}
    \label{fig:lapRL_1}
\end{figure*}

\begin{figure*}
    \centering
    \subfloat[$\beta=0.5$]{\includegraphics[trim={0 0 3.9cm 0}, clip,width=8.4cm,height=8.8cm]{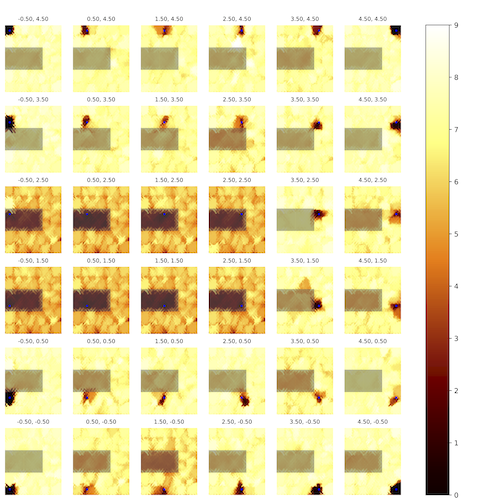}}
    \subfloat[$\beta=1$]{\includegraphics[trim={0 0 1.5cm 0}, clip, width=8.6cm,height=8.8cm]{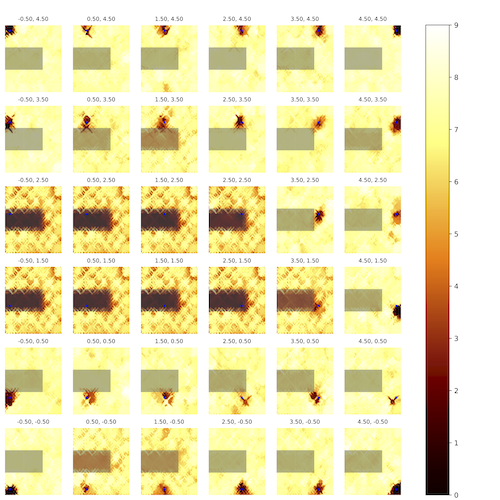}}\\
    \subfloat[$\beta=2$]{\includegraphics[trim={0 0 3.9cm 0}, clip,width=8.4cm,height=8.8cm]{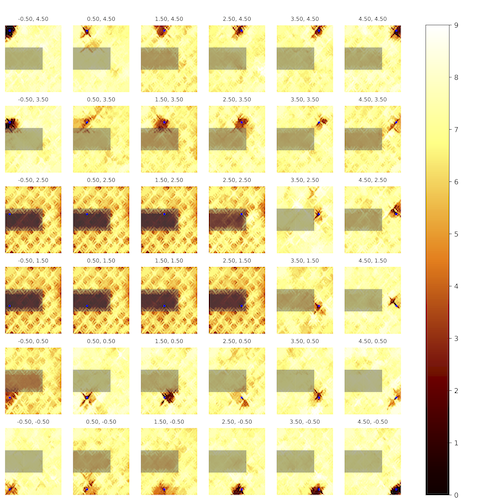}} 
    \subfloat[$\beta=6$]{\includegraphics[trim={0 0 1.5cm 0}, clip, width=8.6cm,height=8.8cm]{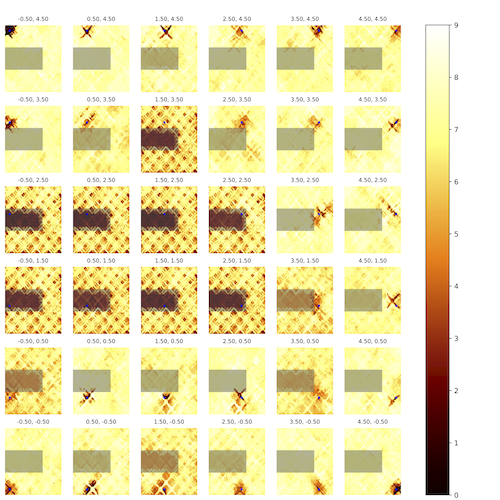}}
    \caption{Laplacian in RL using pixel inputs. Higher $\beta$ better approximates spectral graph drawing objective. $LR=5e-5$}
    \label{fig:lapRL_2}
\end{figure*}

The distance predictor is neural network with $4$ convolution layers with $64$ channels and kernel size of $3$ in each layer with strides $(1, 2, 1, 2)$ followed by $2$ fully connected layers with $128$ units in each layer and the output layer has $32$ units. We used \textit{relu} non-linearity along with batchnorm. The learning rate was set to $5e$-$5$ and Adam \cite{kingma2014method} optimizer. The network was trained for $50$ epochs with $p=2$ and $q=1$. The top-down view of the maze ant is shown in Figure \ref{fig:pixel_top_down_view}. The results with the approach of \cite{wu2018the} are shown in Figures \ref{fig:lapRL_1} and \ref{fig:lapRL_2}. The training setup is the same as described in section \ref{sec:visual_experiment}. We uniformly sample the states in each trajectory (hence, $\lambda$ is approximately $1$).  As $\beta$ is increased (better approximation of the spectral objective objective), the points that are predicted close in almost all of the reference points resembles the body of the ant in the stable position (Figure \ref{fig:pixel_top_down_view}) centered on nearby points. 

\begin{figure*}
    \centering
    \subfloat[$16$]{\includegraphics[trim={0 0 3.9cm 0}, clip,width=8.4cm,height=8.8cm]{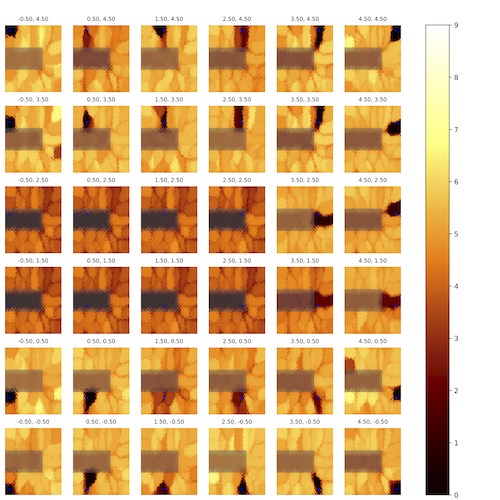}}
    \subfloat[$32$]{\includegraphics[trim={0 0 1.5cm 0}, clip, width=8.6cm,height=8.8cm]{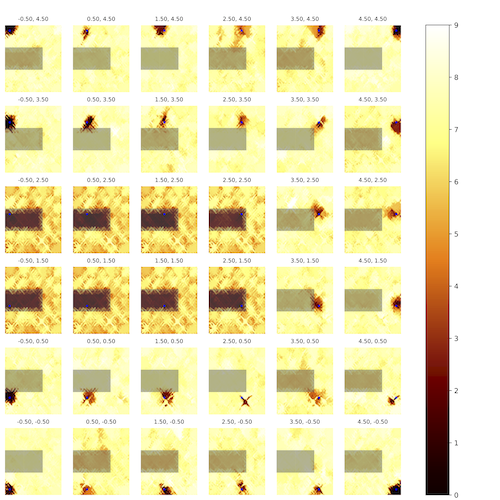}}\\
    \subfloat[$64$]{\includegraphics[trim={0 0 3.9cm 0}, clip,width=8.4cm,height=8.8cm]{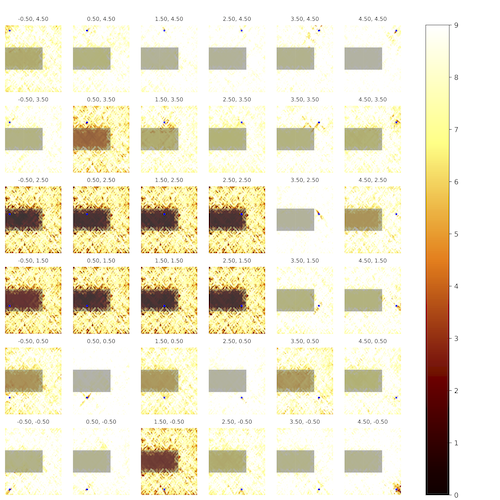}} 
    \subfloat[$128$]{\includegraphics[trim={0 0 1.5cm 0}, clip, width=8.6cm,height=8.8cm]{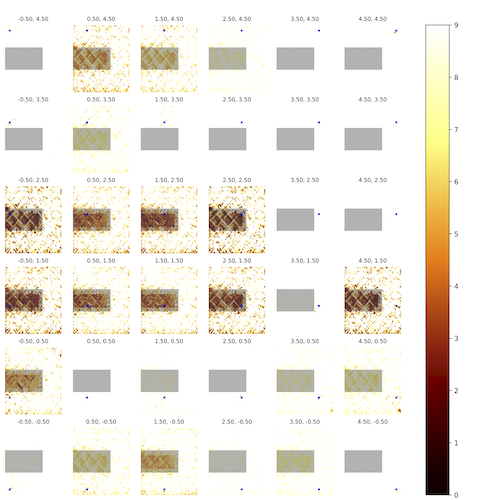}}
    \caption{The effect of size of embeddings using the objective of Laplacian in RL with pixel inputs for $LR=1e-4$ and $\beta=1$. The quality of the approximation of the distance drops after $32$ dimensions.}
    \label{fig:embedding_effect_laprl}
\end{figure*}

\begin{figure*}
    \centering
    \subfloat[$16$]{\includegraphics[trim={0 0 3.9cm 0}, clip,width=8.4cm,height=8.8cm]{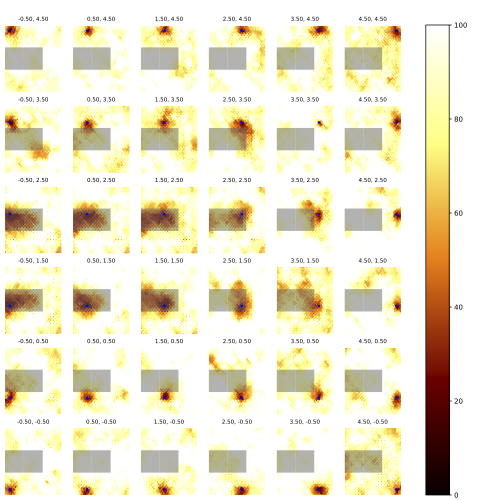}}
    \subfloat[$32$]{\includegraphics[trim={0 0 1.5cm 0}, clip, width=8.6cm,height=8.8cm]{icml2020/images/pixel_task/ours/ours_32_1.png}}\\
    \subfloat[$64$]{\includegraphics[trim={0 0 3.9cm 0}, clip,width=8.4cm,height=8.8cm]{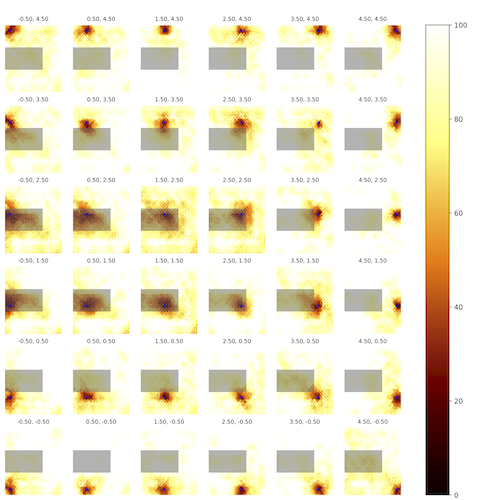}} 
    \subfloat[$128$]{\includegraphics[trim={0 0 1.5cm 0}, clip, width=8.6cm,height=8.8cm]{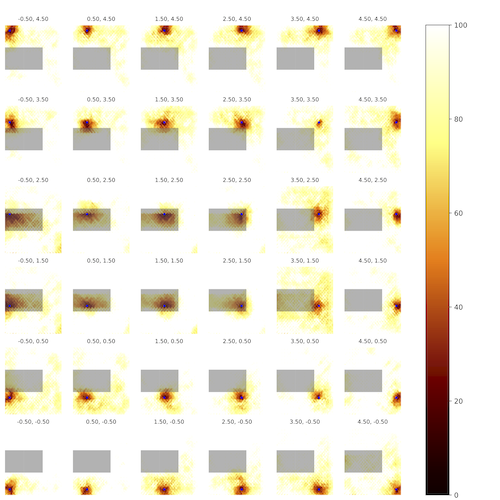}}
    \caption{Increasing the embedding size in our approach improves the distance estimates.}
    \label{fig:embedding_effect_ours}
\end{figure*}

When used in an online setting, the negative sampling in \cite{wu2018the} could be problematic especially when bootstrapping (without arbitrary environment resets). In contrast, our objective function does not require negative sampling and only uses the information present within a trajectory, making it more suitable for online learning. The discussion in \ref{sec:compare_laplacian_pseudo_laplacian} suggests that increasing the embedding dimensions monotonically increases the quality of distance estimates in the scaled spectral embeddings unlike spectral embeddings. We show this phenomenon using the pixel inputs in Figures \ref{fig:embedding_effect_laprl} \cite{wu2018the} and \ref{fig:embedding_effect_ours} (our method).


\begin{figure*}
    \centering{\includegraphics[trim={0 0 1.5cm 0}, clip, width=0.8\textwidth]{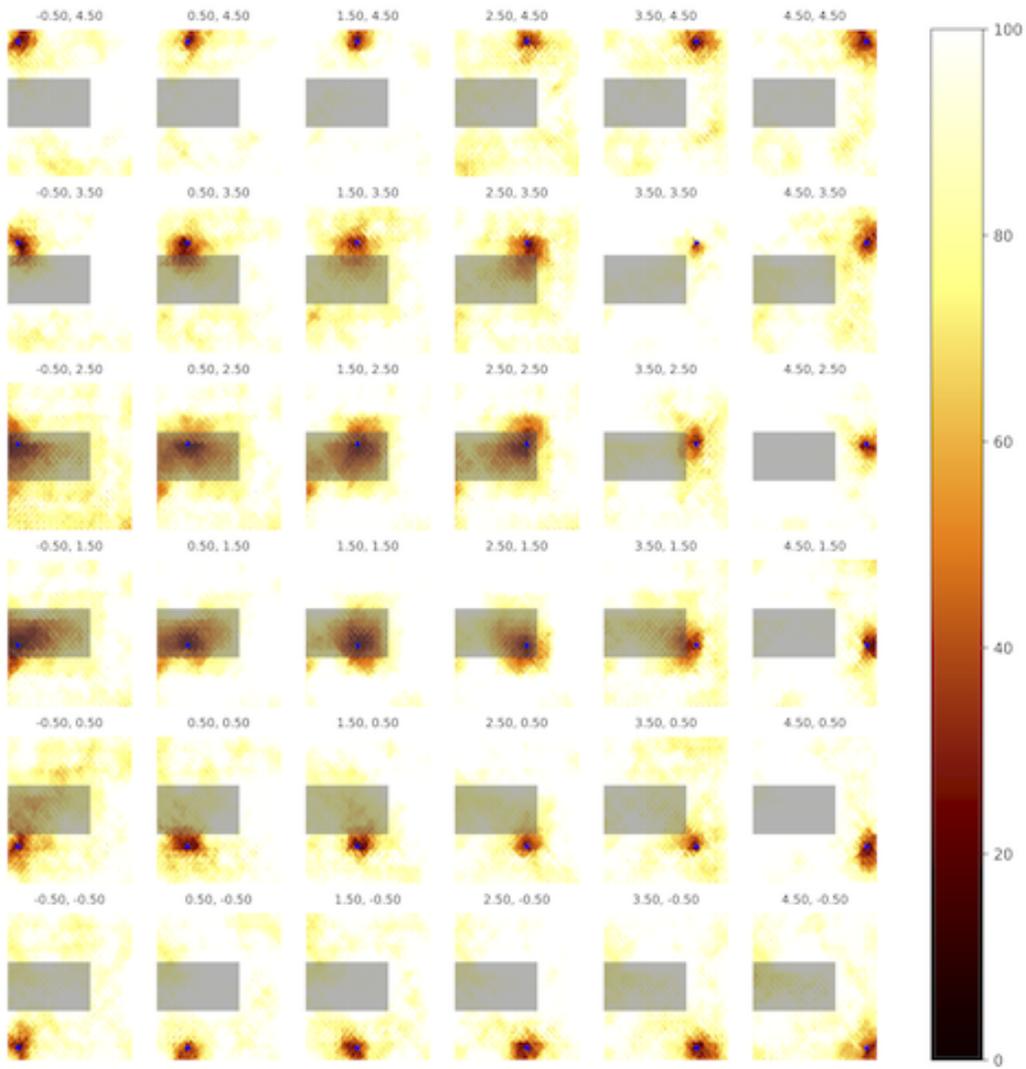}}
    \caption{Full heatmap of our approach using pixel inputs. Our approach does not require negative sampling.}
    \label{fig:ours_pixel}
\end{figure*}